\documentclass[11pt, a4paper]{lumia}
\usepackage[sort&compress]{natbib}
\usepackage{longtable}   
\usepackage{array}      
\usepackage{xltabular}
\usepackage{arydshln}
\usepackage{dashrule}
\usepackage{booktabs}
\usepackage{caption}

\setheadertext{PatRe: A Full-Stage Office Action and Rebuttal Generation Benchmark for Patent Examination}

\renewcommand{\today}{2026-05}

\usepackage{xspace}

\theoremstyle{plain}

\newtheorem*{proposition*}{Proposition}

\theoremstyle{definition}

\theoremstyle{definition}

\def\eqref#1{equation~\ref{#1}}

\usepackage[utf8]{inputenc}     
\usepackage[T1]{fontenc}        
\usepackage{microtype}          

\usepackage{xcolor}             
\usepackage[dvipsnames]{xcolor} 
\usepackage{graphicx}           
\usepackage{tikz}               
\usepackage[edges]{forest}      

\usepackage{hyperref}           
\usepackage{url}                
\usepackage{xurl}               

\usepackage{tabularx}           
\usepackage{array}              
\usepackage{booktabs}           
\usepackage{longtable}          
\usepackage{multirow}           
\usepackage{makecell}           
\usepackage[table]{xcolor}      
\usepackage{colortbl}           
\usepackage{ragged2e}           

\usepackage{amsmath}            
\usepackage{mathtools}          
\usepackage{amsfonts}           
\usepackage{amssymb}            
\usepackage{amsthm}             
\usepackage{nicefrac}           

\usepackage{algorithm}          
\usepackage{algorithmicx}       
\usepackage{algpseudocode}      
\usepackage{listings}           

\usepackage{subcaption}         
\usepackage{caption}            
\usepackage{wrapfig}            
\usepackage[export]{adjustbox}  

\usepackage{soul}               
\usepackage{changes}            
\usepackage{xspace}             
\usepackage[normalem]{ulem}     
\usepackage{CJKutf8}            

\usepackage{enumitem}           
\usepackage{pifont}             

\usepackage[most,skins,theorems]{tcolorbox} 
\usepackage[tikz]{bclogo}       
\usepackage[framemethod=tikz]{mdframed} 

\usepackage{lipsum}             
\usepackage{tocloft}            
\usepackage{fontawesome5}       
\usepackage{afterpage}          
\usepackage{bbding}             
\usepackage{epigraph}           
\usepackage{minitoc}            
\usepackage{multicol}           
\usepackage{textgreek}          

\newcolumntype{P}[1]{>{\RaggedRight\arraybackslash}p{#1}}

\definecolor{darkblue}{rgb}{0, 0, 0.5}
\definecolor{uclablue}{RGB}{39, 116, 174}
\definecolor{bigaired}{RGB}{156, 0, 0}
\definecolor{myblue}{HTML}{598BE7}
\definecolor{mildblue}{RGB}{31,119,180}
\definecolor{sectionblue}{RGB}{70, 130, 180}
\definecolor{methodblue}{RGB}{0, 150, 136}
\definecolor{bgblue}{RGB}{245,243,253}
\definecolor{ttblue}{RGB}{91,194,224}
\definecolor{mygreen}{rgb}{0.64, 0.56, 0.88}
\definecolor{myyellow}{rgb}{0.68, 0.6, 0.1}
\definecolor{fancygreen}{rgb}{0.33, 0.68, 0.20}
\definecolor{salmon}{rgb}{0.94, 0.52, 0.49}
\definecolor{tablegreen}{rgb}{0.82, 0.94, 0.75}
\definecolor{tableblue}{rgb}{0.81, 0.90, 0.94}
\definecolor{tablered}{rgb}{0.97, 0.85, 0.85}
\definecolor{tableorange}{rgb}{0.96, 0.85, 0.81}
\definecolor{myorange}{rgb}{1.0, 0.49, 0.0}
\definecolor{tlgreen}{rgb}{0.33, 0.68, 0.20}
\definecolor{darkgreen}{RGB}{0,100,0}
\definecolor{darkred}{RGB}{200, 0, 0}
\definecolor{lightblue}{RGB}{220,235,250}
\definecolor{customyellow}{HTML}{FFFACD}
\definecolor{refinegreen}{RGB}{0, 128, 75}
\definecolor{scoregreen}{RGB}{34, 139, 34}
\definecolor{hidden-blue}{RGB}{194,232,247}
\definecolor{hidden-black}{RGB}{20,68,106}
\definecolor{yes}{HTML}{C6EFCE}
\definecolor{no}{HTML}{FFC7CE}
\definecolor{partial}{HTML}{FFEB9C}
\definecolor{external}{HTML}{D9E1F2}
\definecolor{hdr}{HTML}{F2F2F2}
\definecolor{GRPOrow}{gray}{0.96}
\definecolor{FlowRLrow}{RGB}{225,236,255}
\definecolor{FlowBlue}{RGB}{80,120,210}
\definecolor{GRPOGray}{gray}{0.35}

\hypersetup{
    colorlinks=true, 
    citecolor=uclablue, 
    linkcolor=bigaired,
    urlcolor=darkblue
}

\setlist[itemize]{leftmargin=20pt, noitemsep, topsep=0pt}

\NewDocumentCommand{\kaiyan}{mO{}}{\textcolor{purple}{\textsuperscript{\textit{kaiyan}}\textsf{\textbf{\small[#1]}}}}
\NewDocumentCommand{\yuxin}{mO{}}{\textcolor{cyan}{\textsuperscript{\textit{yuxin}}\textsf{\textbf{\small[#1]}}}}
\NewDocumentCommand{\bx}{mO{}}{\textcolor{green}{\textsuperscript{\textit{bx}}\textsf{\textbf{\small[#1]}}}}
\NewDocumentCommand{\at}{mO{}}{\textcolor{red}{\textsuperscript{\textit{AT}}\textsf{\textbf{\small[#1]}}}}
\NewDocumentCommand{\re}{mO{}}{\textcolor{blue}{\textsuperscript{\textit{RE}}\textsf{\textbf{\small[#1]}}}}
\NewDocumentCommand{\ybsun}{mO{}}{\textcolor{magenta}{\textsuperscript{\textit{youbang}}\textsf{\textbf{\small[#1]}}}}
\NewDocumentCommand{\runze}{mO{}}{\textcolor{orange}{\textsuperscript{\textit{runze}}\textsf{\textbf{\small[#1]}}}}
\NewDocumentCommand{\add}{mO{}}{\textcolor{darkgreen}{\textsuperscript{\textit{Maybe Consider Discuss}}\textsf{\textbf{[#1]}}}}

\newcommand{\cmark}{\textcolor{darkgreen}{\boldmath$\checkmark$}}
\newcommand{\xmark}{\textcolor{darkred}{\boldmath$\times$}}

\newenvironment{itemize*}%
 {\leftmargini=10pt\begin{itemize}%
  \setlength{\itemsep}{0pt}%
  \setlength{\parskip}{0pt}%
  }%
 {\end{itemize}}

\newenvironment{enumerate*}%
 {\begin{enumerate}%
  \setlength{\itemsep}{0pt}%
  \setlength{\parskip}{0pt}}%
 {\end{enumerate}}

\newcommand{\cellstatus}[1]{%
  \begingroup
  \StrTrim{#1}[\statusval]%
  \IfStrEq{\statusval}{Yes}{\cellcolor{yes}\cmark}{}%
  \IfStrEq{\statusval}{No}{\cellcolor{no}\xmark}{}%
  \IfBeginWith{\statusval}{Yes (}{\cellcolor{yes}\cmark~\textit{\statusval\unskip}}{}%
  \IfStrEq{\statusval}{Partial}{\cellcolor{partial}\textbf{Partial}}{}%
  \IfStrEq{\statusval}{External}{\cellcolor{external}\textbf{External}}{}%
  \endgroup
}

\newtcolorbox{myboxi}[1][]{
  breakable,
  title=#1,
  colback=red!5,
  colbacktitle=red!5,
  coltitle=black,
  fonttitle=\bfseries,
  bottomrule=0pt,
  toprule=0pt,
  leftrule=2pt,
  rightrule=2pt,
  titlerule=0pt,
  arc=0pt,
  outer arc=0pt,
  colframe=red,
}

\newtcolorbox{myboxnote}[1][]{
  breakable,
  title=#1,
  colback=orange!0,
  colbacktitle=orange!0,
  coltitle=black,
  fonttitle=\bfseries,
  bottomrule=0pt,
  toprule=0pt,
  leftrule=2pt,
  rightrule=2pt,
  titlerule=0pt,
  arc=0pt,
  outer arc=0pt,
  colframe=orange,
}

\newtcolorbox{myboxii}[1][]{
  breakable,
  freelance,
  title=#1,
  colback=white,
  colbacktitle=white,
  coltitle=black,
  fonttitle=\bfseries,
  bottomrule=0pt,
  boxrule=0pt,
  colframe=white,
  overlay unbroken and first={
  \draw[red!75!black,line width=3pt]
    ([xshift=5pt]frame.north west) -- 
    (frame.north west) -- 
    (frame.south west);
  \draw[red!75!black,line width=3pt]
    ([xshift=-5pt]frame.north east) -- 
    (frame.north east) -- 
    (frame.south east);
  },
  overlay unbroken app={
  \draw[red!75!black,line width=3pt,line cap=rect]
    (frame.south west) -- 
    ([xshift=5pt]frame.south west);
  \draw[red!75!black,line width=3pt,line cap=rect]
    (frame.south east) -- 
    ([xshift=-5pt]frame.south east);
  },
  overlay middle and last={
  \draw[red!75!black,line width=3pt]
    (frame.north west) -- 
    (frame.south west);
  \draw[red!75!black,line width=3pt]
    (frame.north east) -- 
    (frame.south east);
  },
  overlay last app={
  \draw[red!75!black,line width=3pt,line cap=rect]
    (frame.south west) --
    ([xshift=5pt]frame.south west);
  \draw[red!75!black,line width=3pt,line cap=rect]
    (frame.south east) --
    ([xshift=-5pt]frame.south east);
  },
}

\tcbset{
  takeawaysbox/.style={
    title=Takeaways,
    colback=lightblue!80,
    colframe=black,
    fonttitle=\bfseries\small,
    coltitle=white,
    colbacktitle=black,
    enhanced,
    attach boxed title to top left={xshift=2.5mm,yshift=-2.5mm},
    boxed title style={rounded corners, size=small, colframe=black, colback=black},
    width=\linewidth,
    arc=3.5mm
  }
}

\mdfdefinestyle{mystyle}{%
  rightline=true,
  innerleftmargin=10,
  innerrightmargin=10,
  outerlinewidth=3pt,
  topline=false,
  rightline=true,
  bottomline=false,
  skipabove=\topsep,
  skipbelow=\topsep
}

\tikzset{%
    every node/.style={font=\tiny},
    parent/.style =          {align=center,text width=2cm,rounded corners=3pt, line width=0.3mm, fill=gray!10,draw=gray!80},
    child/.style =           {align=center,text width=2.0cm,rounded corners=3pt, fill=blue!10,draw=blue!80,line width=0.3mm},
    grandchild/.style =      {align=center,text width=2cm,rounded corners=3pt},
    greatgrandchild/.style = {align=center,text width=1.5cm,rounded corners=3pt},
    greatgrandchild2/.style = {align=center,text width=1.5cm,rounded corners=3pt},    
    referenceblock/.style =  {align=center,text width=1.5cm,rounded corners=2pt},
    pretrain/.style =           {align=center,text width=2.0cm,rounded corners=3pt, fill=blue!10,draw=blue!80,line width=0.3mm},   
    pretrain_work/.style =           {align=center, text width=8.5cm,rounded corners=3pt, fill=blue!10,draw=blue!0,line width=0.3mm},  
    template/.style =           {align=center,text width=2.0cm,rounded corners=3pt, fill=red!10,draw=red!80,line width=0.3mm},   
    template_work/.style =           {align=center,text width=8.5cm,rounded corners=3pt, fill=red!10,draw=red!0,line width=0.3mm},    
    answer/.style =           {align=center,text width=2.0cm,rounded corners=3pt, fill= cyan!10,draw= cyan!80,line width=0.3mm},   
    answer_work/.style =           {align=center,text width=8.5cm,rounded corners=3pt, fill= cyan!10,draw= cyan!0,line width=0.3mm},      
    multiple/.style =           {align=center,text width=2.0cm,rounded corners=3pt, fill= orange!10,draw= orange!80,line width=0.3mm},   
    multiple_work/.style =           {align=center,text width=8.5cm,rounded corners=3pt, fill= orange!10,draw= orange!0,line width=0.3mm},        
    tuning/.style =           {align=center,text width=2.0cm,rounded corners=3pt, fill= magenta!10,draw= magenta!80,line width=0.3mm},   
    tuning_work/.style =           {align=center,text width=8.5cm,rounded corners=3pt, fill= magenta!10,draw= magenta!0,line width=0.3mm},          
}

\newcommand{\lstbg}[3][0pt]{{\fboxsep#1\colorbox{#2}{\strut #3}}}

\lstdefinelanguage{diff}{
  basicstyle=\ttfamily\small,
  morecomment=[f][\lstbg{red!20}]-,
  morecomment=[f][\lstbg{green!20}]+,
}

\lstdefinelanguage{diffpython}{
  language=diff,
  morekeywords={def, if, else, for, while, return, import, from, as, class, with, try, except, finally, raise, lambda, and, or, not, in, is, None, True, False},
  morecomment=[l]{\#},
  morestring=[b]",
  morestring=[b]',
}

\usepackage[most,skins,theorems]{tcolorbox}
\newcommand{\cYes}{\textcolor{green!60!black}{\textbf{Yes}}} 
\newcommand{\cNo}{No} 
\newcommand{\cPart}{Partial}
\definecolor{over-enforcement}{HTML}{8BABD3}
\definecolor{dual failure}{HTML}{E39889}

\newtcolorbox{promptbox}[2][]{
  colback=gray!5,       
  colframe=black!70,    
  title=\textbf{#2},     
  fonttitle=\bfseries\small,
  fontupper=\footnotesize\ttfamily, 
  left=3pt, right=3pt, top=3pt, bottom=3pt, 
  boxrule=0.8pt,
  sharp corners,         
  #1
}

\title{PatRe: A Full-Stage Office Action and Rebuttal Generation Benchmark for Patent Examination}

\author{%
    \scriptsize
    Qiyao Wang$^{1,2,*}$, Xinyi Chen$^{3,*}$, 
    Longze Chen$^{1,2}$, Hongbo Wang$^{3}$,
    Hamid Alinejad-Rokny$^{4}$,
    Yuan Lin$^{3\dagger}$, Min Yang$^{1,5\dagger}$ \\
    $^1$~\raisebox{-0.5ex}{\includegraphics[height=2.9ex]{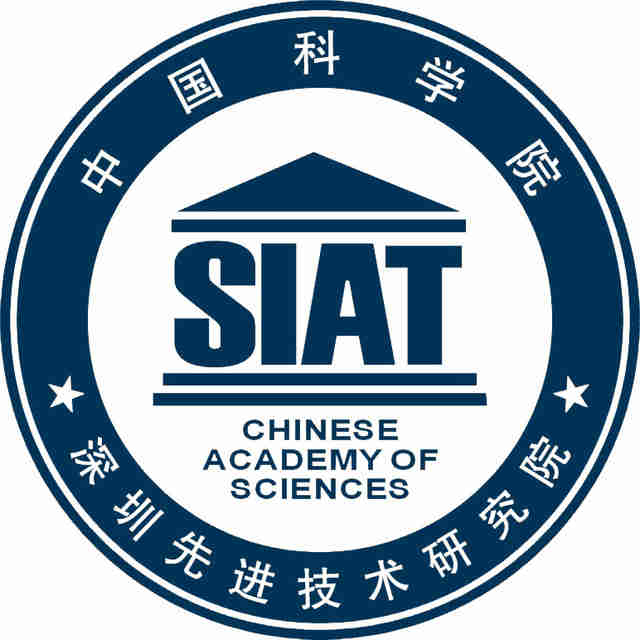}}~Shenzhen Institute of Advanced Technology, Chinese Academy of Sciences \quad
    $^2$~\raisebox{-0.5ex}{\includegraphics[height=2.9ex]{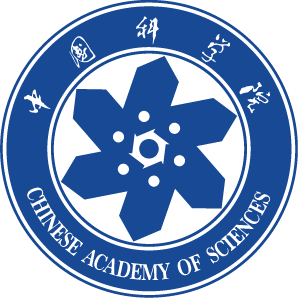}}~University of Chinese Academy of Sciences\\
    $^3$~\raisebox{-0.5ex}{\includegraphics[height=2.9ex]{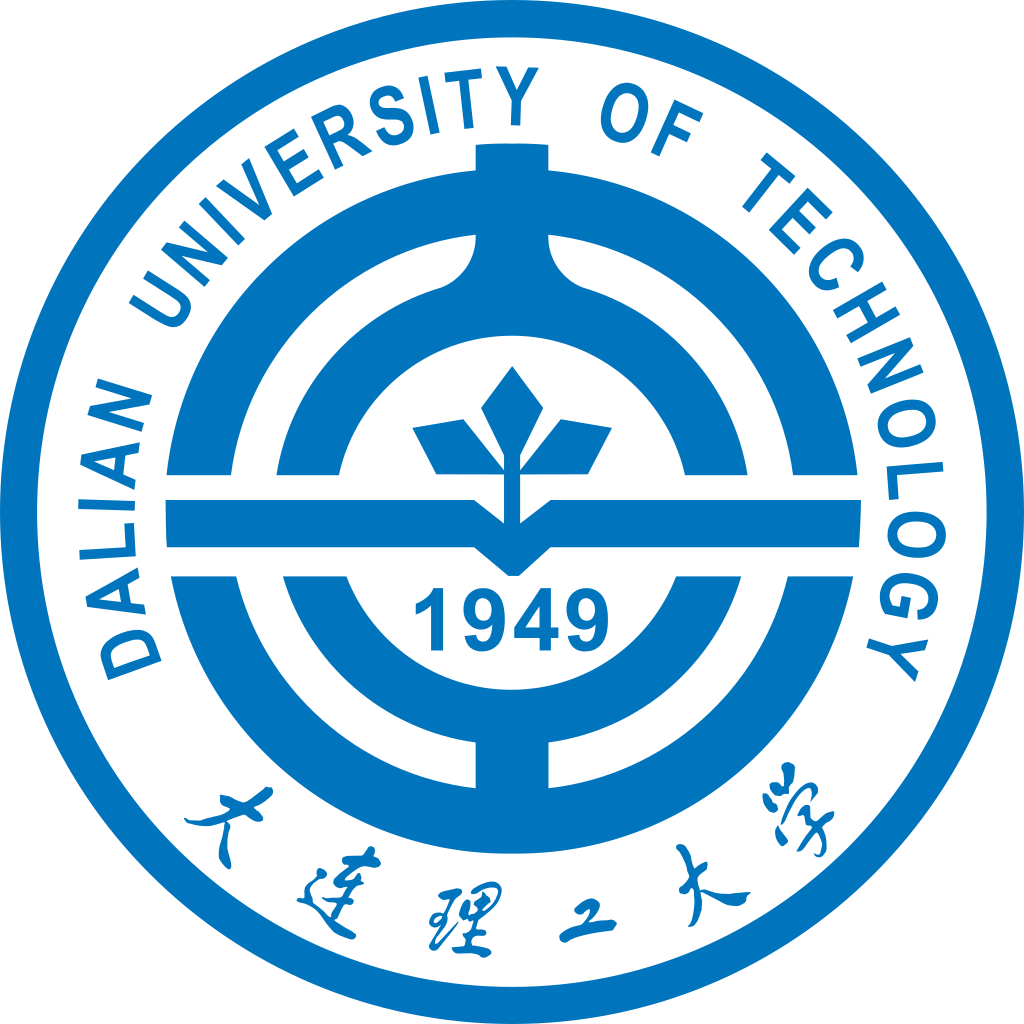}}~Dalian University of Technology\quad
    $^4$~\raisebox{-0.5ex}{\includegraphics[height=2.9ex]{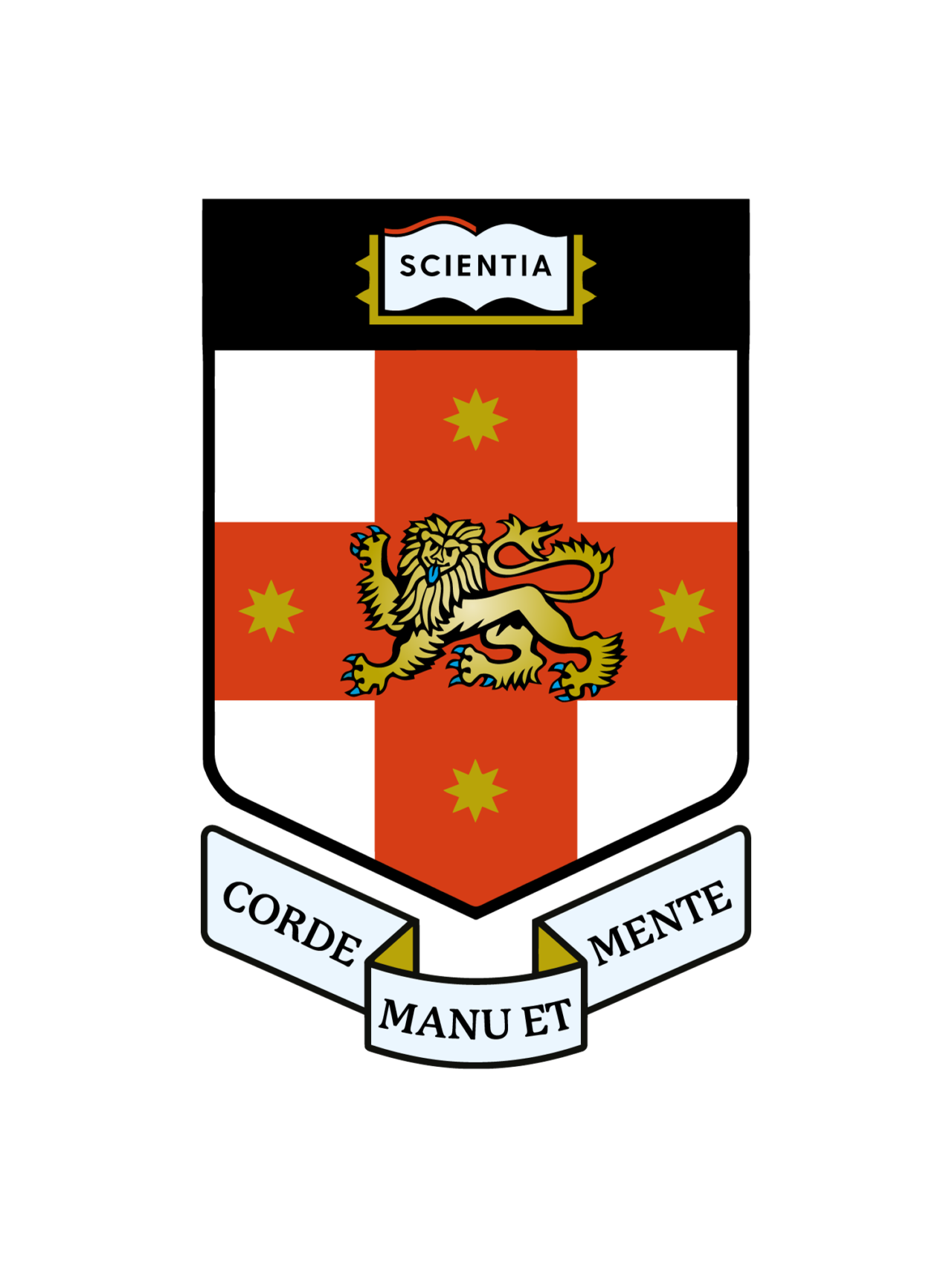}}~UNSW Sydney\quad
    $^5$~\raisebox{-0.5ex}{\includegraphics[height=2.9ex]{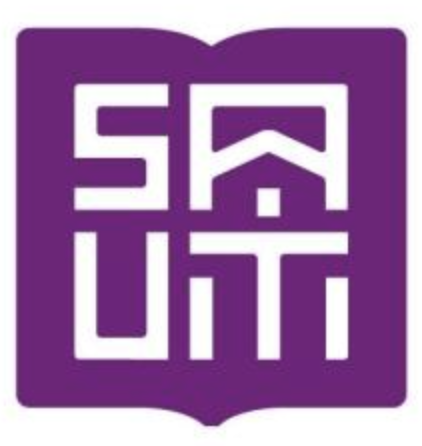}}~Shenzhen University of Advanced Technology \\
    \faEnvelope[regular]~\texttt{wangqiyao25@mails.ucas.ac.cn}  \quad
    \faEnvelope[regular]~\texttt{zhlin@dlut.edu.cn}  \quad
    \faEnvelope[regular]~\texttt{min.yang@siat.ac.cn}  \\
    \faHome~\href{https://patre.wangqiyao.me/}{Website} \quad \faGithub~\href{https://github.com/AIforIP/PatRe}{PatRe} \quad
    $^*$ Equal Contribution. \quad
    $^\dagger$ Corresponding Authors. 
}

\begin{abstract}
Patent examination is a complex, multi-stage process requiring both technical expertise and legal reasoning, increasingly challenged by rising application volumes. 
Prior benchmarks predominantly view patent examination as discriminative classification or static extraction, failing to capture its inherently interactive and iterative nature, similar to the peer review and rebuttal process in academic publishing.
In this paper, we introduce \textbf{PatRe}, the first benchmark that models the full patent examination lifecycle, including Office Action generation and applicant rebuttal. 
PatRe comprises 480 real-world cases and supports both oracle and retrieval-simulated evaluation settings.
Our benchmark reframes patent examination as a dynamic, multi-turn process of justification and response.
Extensive experiments across various LLMs reveal critical insights into model performance, including differences between proprietary and open-source models, as well as task asymmetries between examiner analysis and applicant-side rebuttal.
These findings highlight both the potential and current limitations of LLMs in modeling complex, real-world legal reasoning and technical novelty judgment in patent examination.
We release our code and dataset to facilitate future research on patent examination modeling.
\end{abstract}

\begin{document}

\maketitle

\section{Introduction}

Patent examination is a critical process that ensures applications are sufficiently novel, non-obvious, useful, and meet statutory requirements to be granted.
With the rapid growth of patent applications across various fields and the rigorous processes in different jurisdictions’ Intellectual Property (IP) Offices, patent examiners face increasing pressure.
For example, in 2025, the United States Patent and Trademark Office (USPTO) received 475,223 patent applications, with a backlog of 837,928 unexamined applications and a first-action pendency of 20.5 months.
With advancements in large language models (LLMs)~\citep{hurst2024gpt,liu2024deepseek}, \citet{knappich-etal-2025-pap2pat} and \citet{wang2024autopatent} develop LLM-based and agent-based approaches to automatically generate patent documents, which exacerbate this issue and place greater demands on examiners, requiring stricter review.
These issues also stem from the complexity of patent examination, which requires examiners to be not only well-versed in the relevant technical field but also knowledgeable about patent law. 
The examiner must carefully review the new patent application and use search tools for the prior art to determine whether it is useful, non-obvious, statutory, and novel as outlined in the Manual of Patent Examining Procedure (MPEP) \citep{uspto2020mpep}. 

Researchers have made significant efforts to leverage AI in assisting the patent examination process. 
HUPD~\citep{suzgun2023harvard} first introduce the discriminative \textit{Acceptance Prediction} task, a binary classification that inputs a patent’s abstract or claims and uses BERT-like models to predict acceptance or rejection. 
Beyond coarse-grained classification, PANORAMA~\citep{lim2025panorama} focuses on more fine-grained classification of rejection reasons, introducing the NOC4PC task, which is aligned with legal basis codes, particularly §102 and § 103.
It also introduces the PAR4PC task, which assesses conflicts with the novelty of prior arts.
All these examination-related tasks adopt a discriminative manner, lacking interpretability and detailed analysis for rejection or grant decisions.

In the patent examination practice, an Office Action (OA) is not a one-time event. 
Applicants can submit a rebuttal to the examiner’s OA in hopes of obtaining a grant until a final decision is reached, similar to the discussion and rebuttal process in peer review of academic papers~\citep{zhang2025re,Li2025AutomaticPR}. 
However, prior work focus only on reviewing the initial version of a patent application, overlooking the multi-turn interaction between the examiner and applicant and the evolution of subsequent patent versions. 
Additionally, all these works rely on acceptance or statute accuracy as the metric, lacking a fine-grained analysis of the correctness of the examination suggestions.

\begin{figure}[!t]
\includegraphics[width=\linewidth]{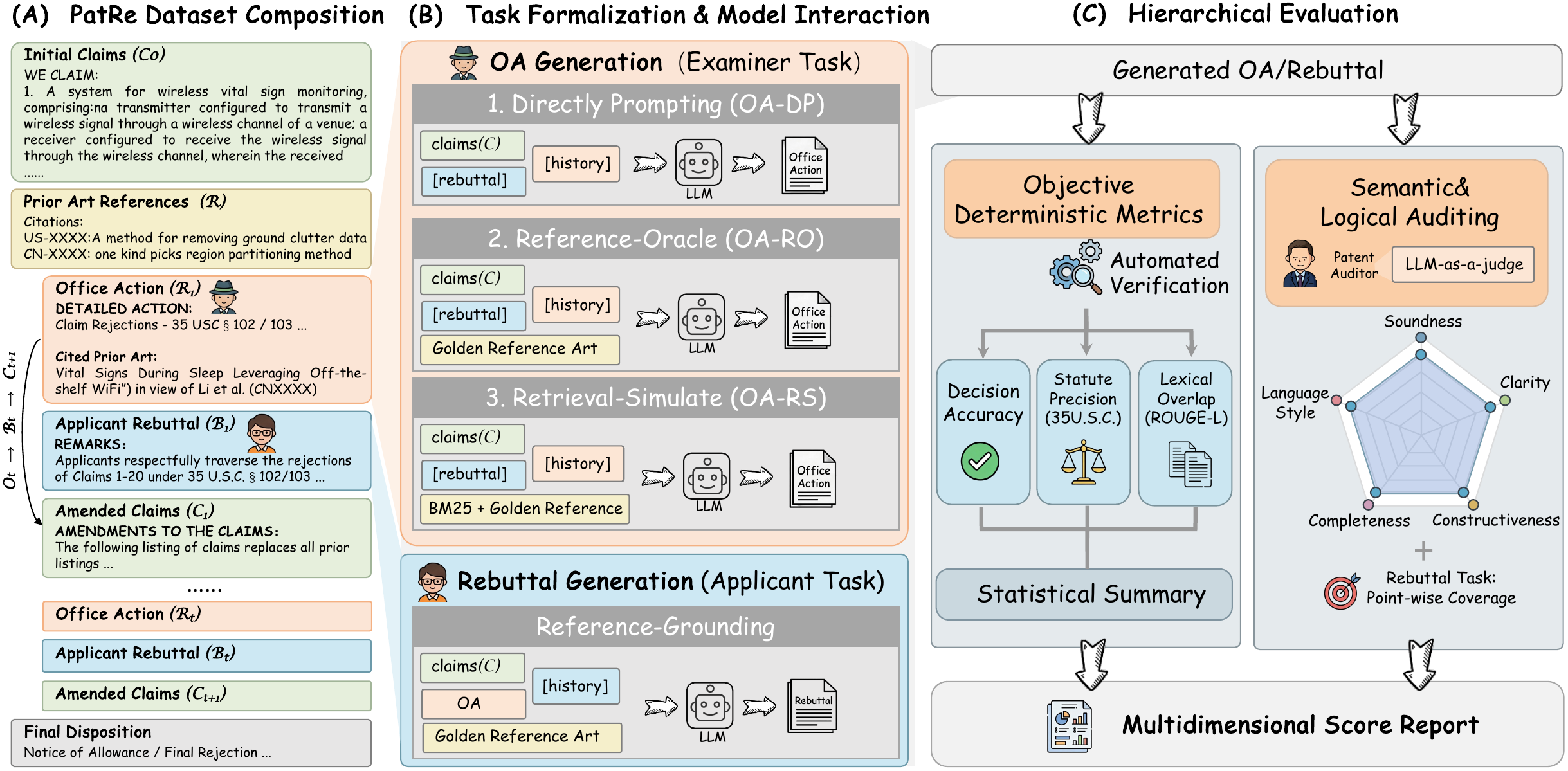} 
    \caption{The overall framework of the PatRe benchmark.}
    \label{fig:framework_overview}
\end{figure}

In this work, we focus on the entire patent examination lifecycle and introduce \textbf{PatRe}, the first full-stage benchmark of \textbf{\underline{Pat}}ent Office Actions and \textbf{\underline{Re}}buttals generation, as illustrated in Figure~\ref{fig:framework_overview}. It primarily includes two types of tasks: 
\textbf{(I)} Office Action (OA) Generation, which requires the model to produce formal examination reports by analyzing patent claims against potential prior art. 
Beyond direct prompting, we further distinguish OA generation into two settings: \textit{Reference-Oracle Generation}, which provides oracle citations to assess the model’s upper-bound capability under ideal evidence conditions, and \textit{Retrieval-Simulate Generation}, which simulates real-world scenarios by additionally supplying prior art retrieved via BM25. 
In retrieval-simulated setting, the model must first identify the relevant prior art and assess its relevance before generating its response.
\textbf{(II)} Rebuttal Generation, which assesses the model's capacity to simulate applicant responses. 
Given an examiner's OA, the model must generate legal and technical remarks to contest specific rejection grounds and provide persuasive arguments to overcome the cited prior art, focusing on the logical consistency and legal validity of the defense.
\textbf{Our main contributions are as follows:}

\begin{itemize}
    \item We introduce the first full-stage patent examination benchmark, \textbf{PatRe}, which focuses on the entire lifecycle of multi-turn Office Action and rebuttal generation. It contains 480 recent patent examination records and covers diverse IPC fields and legal attributes, including final Office Action decisions, intermediate rejection types, and examiner-cited reference patents.
    \item Moving beyond discriminative classification and static extraction, we view patent examination as a dynamic process of justification between the examiner and the applicant. Notably, given the novelty assessment requirements in Office Action generation, we evaluate LLMs under varying levels of cited reference exposure and noise, aligning with realistic examination procedures.
    \item We conduct extensive experiments on a range of LLMs, providing in-depth insights from legal and domain-specific perspectives, including the gap between proprietary and open-source models, the asymmetry between proactive examination (OA) and reactive advocacy (rebuttal), and broader analytical findings.
\end{itemize}

\section{Related Work}

\begin{table*}[!h]
\small
\setlength{\tabcolsep}{6pt}
\resizebox{\textwidth}{!}{\begin{tabular}{lllccccc}
\toprule
\textbf{Category} & \textbf{Dataset} & \textbf{Year} & \textbf{Task} & \textbf{Statute} & \textbf{Evolution} & \textbf{Adversarial} & \textbf{Full-stage} \\
\midrule
\multirow{6}{*}{\makecell[l]{Classification\\\& Extraction}} 
    & HUPD \citep{suzgun2023harvard} & 2023 & Discriminative & \cNo & \cNo & \cNo & \cNo \\
    & IPEval \citep{Wang2024IPEvalAB} & 2024 & Discriminative & \cYes & \cNo & \cNo & \cNo \\
    & IPBench \citep{wang2025ipbench} & 2025 & Discriminative & \cNo & \cNo & \cNo & \cNo \\
    & PILOT-Bench \citep{jang2025pilot} & 2025 & Discriminative & \cYes & \cNo & \cNo & \cNo \\
    & PANORAMA \citep{lim2025panorama} & 2025 & Discriminative & \cYes & \cNo & \cNo & \cNo \\
    & PEDANTIC \citep{knappich2025pedantic} & 2025 & Discriminative & \cPart & \cNo & \cNo & \cNo \\
\midrule
\multirow{3}{*}{\makecell[l]{Revision\\\& Drafting}} 
    & MOZIP \citep{ni2024mozip} & 2024 & Discriminative & \cNo & \cNo & \cNo & \cNo \\
    & PatentEdits \citep{lee2024patentedits} & 2024 & Generative & \cNo & \cYes & \cNo & \cPart \\
    & Patent-CR \citep{jiang2025patent} & 2025 & Generative & \cNo & \cYes & \cNo & \cPart \\
\midrule
\textbf{Ours} & \textbf{PatRe} & \textbf{2026} & \textbf{Generative} & \textbf{\cYes} & \textbf{\cYes} & \textbf{\cYes} & \textbf{\cYes} \\
\bottomrule
\end{tabular}}
\caption{Comparison of {PatRe} with related patent datasets and benchmarks. \textbf{Statute}: Explicit legal basis. \textbf{Evolution}: Claim versioning. \textbf{Adversarial}: Multi-turn interaction.}
\label{tab:comparison}
\end{table*}

\paragraph{Binary Patent Classification and Static Justification Extraction.}
Early work primarily view patent examination as a classification task. 
HUPD \citep{suzgun2023harvard} utilize BERT-based models for acceptance prediction, while IPBench~\citep{wang2025ipbench} extend this to the modern LLMs.
Additionally, PILOT-Bench~\citep{jang2025pilot} align the patent board decisions with IRAC framework.
However, these works remain restricted to post-hoc classification and treat legal reasoning as a static annotation problem.
They fail to capture the proactive drafting logic and lack the modeling of multi-turn OA generation.
To move toward explainable examination, recent studies have targeted specific legal statutes. 
PANORAMA \citep{lim2025panorama} introduced rejection reason identification, while PEDANTIC \citep{knappich2025pedantic} focused on 35 U.S.C. 112~(b) by performing justification extraction from Office Actions.
Although these provide granular insights, they remain single-stage, static analyses. 
They do not account for the generative complexity of a full OA, nor do they support the multi-turn generation of legal justifications across the iterative dialogue between examiners and applicants.

\paragraph{Claim Revision and Patent Drafting.} 
Another research direction of patent examination investigates how patent claims evolve over time. 
PatentEdits \citep{lee2024patentedits} and Patent-CR \citep{jiang2025patent} aligned initial applications with granted versions to study claim revisions.
While these datasets capture the results of the prosecution process, they primarily focus on static version alignment, omitting the explicit examiner-applicant discussion that fundamentally drives these revisions.
Additionally, Pap2Pat \citep{knappich-etal-2025-pap2pat} and AutoPatent \citep{wang2024autopatent} explored synthesizing patent documents, potentially increasing patent applications and intensifying the need for efficient patent examination.
Researchers have developed benchmarks and methods for academic peer review~\citep{Jin2024AgentReviewEP,Li2025AutomaticPR} and rebuttals~\citep{zhang2025re,Ma2026Paper2RebuttalAM,he2026dancing} to simulate iterative scientific communication. 
However, the patent examination process lacks such benchmarks, which demands stricter adherence to laws like the MPEP~\citep{uspto2020mpep}. 
As shown in Table~\ref{tab:comparison}, our PatRe benchmark bridges this gap by providing the first full-stage benchmark for multi-turn generation of OAs and rebuttals, enabling modeling of the entire examination lifecycle.

\section{PatRe Benchmark}
\subsection{Task Taxonomy and Formalization}

As shown in Figure~\ref{fig:framework_overview}, we conceptualize the patent examination process as a multi-turn strategic interaction between an Examiner ($E$) and an Applicant ($A$). 
Where $\mathcal{D}$ denotes the complete examination history of a given patent, and $|\mathcal{D}|$ represents the number of discussion rounds for current patent.
At each turn $t$, the process is grounded in the current version of claims $\mathcal{C}_t$ and the provided prior art $\mathcal{R}$. 
The examiner first issues an Office Action~(OA) $\mathcal{O}_t$ by evaluating $\mathcal{C}_t$ against $\mathcal{R}$ to identify legal and technical defects. 
Subsequently, the applicant responds with a rebuttal $\mathcal{B}_t$, which provides arguments to contest the rejections or justifies further amendments to the claims $\mathcal{C}_t \rightarrow \mathcal{C}_{t+1}$.
We simulate this entire process by introducing two primary types of tasks, as detailed below:

\paragraph{Task 1: Office Action Generation.} The objective of OA generation is to evaluate model's ability to simulate the examiner's decision-making process. 
Given the current version of claims $\mathcal{C}_t$ and the potentially preceding rebuttal $\mathcal{B}_{t-1}~({\text{when }t>1 }) $, the model is instructed to generate a formal OA $\mathcal{O}$.
We formalize this under three settings with varying levels of information guidance:
\begin{itemize}
    \item \textit{Directly Prompting}~\textbf{(OA-DP)}: We leverage a zero-shot prompting setting to instruct the model generate the Office Action $\mathcal{O}$ by relying solely on its internal parameters and pre-trained knowledge, without access to any specific external prior art.
    \item \textit{Reference-Oracle Generation}~\textbf{(OA-RO)}: We provide the model with an oracle reference set $\mathcal{R}_{\text{oracle}}$, consisting of the ground-truth references cited by the examiner as well as the references cited by the applicant and considered during examination. 
    The model must autonomously select the most revelant references from $\mathcal{R}_{oracle}$ to construct legal justifications for the Office Action $\mathcal{O}$. 
    This subtask evaluates the model’s performance under the most comprehensive information setting.
    \item \textit{Retrieval-Simulated Generation}~\textbf{(OA-RS)}: To simulate a realistic patent examination scenario with retrieval environment, the model is provided with a noisy candidate pool $\mathcal{R}_{noise}$, consisting of top-$k$ references retrieved via BM25 alongside randomly sampled ground-truth references. 
    The model must distinguish pertinent prior art from irrelevant noise to generate the Office Action $\mathcal{O}$.
\end{itemize}

\paragraph{Task 2: Applicant Rebuttal Generation.} This task simulates the responsive phase of patent examination. 
Given the current Office Action $\mathcal{O}_t$ and the associated prior art $\mathcal{R}$ at turn $t$, the model must generate a rebuttal $\mathcal{B}_t$. 
Unlike procedural legal filings, we focus on the substantive argumentation required to overcome the examiner's objections. 
Formally, we model this as $P(\mathcal{B}_t | \mathcal{C}_t, \mathcal{O}_t, \mathcal{R})$, which requires the model to perform a tripartite alignment:
\textbf{(\textit{i})} grounding legal arguments in the specific rejection grounds of $\mathcal{O}_t$, 
\textbf{(\textit{ii})} contrasting the technical features of $\mathcal{C}_t$ against $\mathcal{R}$, and 
\textbf{(\textit{iii})} maintaining logical consistency with the intended scope of the invention.

These tasks collectively establish the PatRe benchmark for evaluating multi-dimensional technical and legal reasoning within patent domain. 
Where PatRe-OA challenges the model's statutory interpretability by requiring the mapping of claim features to prior art disclosures under legal constraints.
Conversely, PatRe-Rebuttal assesses  adversariality through the model's proficiency in synthesizing counter-arguments that adhere to both the technical scope and the MPEP legal framework. 

\subsection{Evaluation Metric Design}
To provide a comprehensive assessment of the generated Office Action and rebuttal documents, we establish a hierarchical evaluation framework that moves beyond surface-level linguistic similarity to capture legal and technical nuances, which including two levels: (I)~Deterministic metrics for objective verification; (II)~LLM-as-a-Judge metrics for deep semantic and logical auditing.

\paragraph{I: Objective Deterministic Metrics.}
To ensure the factual correctness of the generated OA and rebuttal, we implement an objective metric suite that measure alignment with expert-verified label. 
\textbf{(1) Statutory and Decision Alignment}, the accuracy of the legal basis and final decision. 
We compute \textit{Decision Accuracy} as a binary indicator of whether the predicted Office Action decision matches the label. 
Where more fine-grained \textit{Statute Precision} measures the precision of the invoked 35 U.S.C. statutes, \textit{i.e.} $\text{Statute Precision} = \frac{|S_{\text{pred}} \cap S_{\text{gt}}|}{|S_{\text{pred}}|}$,
where $S_{\text{pred}}$ and $S_{\text{gt}}$ denote the sets of statutes cited in the generated and ground-truth legal documents, respectively.
\textbf{(2) Lexical Overlap}, which adopt {Rouge-L}~\citep{lin-2004-rouge} to measure the sequential alignment between generated and ground-truth texts.

\paragraph{II: Semantic and Logical Auditing (LLM-as-a-Judge).}
To evaluate the semantic and logical quality of generated OA and rebuttal, we employ \texttt{Gemini-3.1-Flash-Lite} as a patent auditor, following~\citep{wang2025ipbench}. 
Each document is scored on a 1-10 scale across five dimensions: 
\textbf{(1) Soundness}, which evaluates the technical and legal soundness of the generated texts;
\textbf{(2) Clarity}, which focuses on the legal readability, logical coherence, and specific structure;
\textbf{(3) Constructiveness}, which emphasizes the actionability of the model response;
In OA generation, it measures the usefulness of examiner guidance; in rebuttal generation, it reflects the strength of counterarguments.
\textbf{(4) Completeness}, which focuses on the utility of the feedback;
\textbf{(5) Language Style}, which focuses on adherence to the legal style and procedural conventions of Office Action and rebuttal drafting. 
Especially for rebuttal generation task, we introduce the \textbf{Point-wise Coverage}, which evaluates the responsive rate to atomic OA rejection points, providing a semantic measure of defense thoroughness.

\subsection{Data Collection and Processing}
To construct the PatRe benchmark, we develop a reproducible data collection pipeline to extract the longitudinal examination history of patents from the USPTO public database.
Unlike prior datasets, which typically capture final version of granted patents, we focus on reconstructing the complete trajectory of a patent application by recording the full history sequence of examiner-applicant interactions. 
For each patent record, we collect the full-stage correspondence starting from the initial filing, including the verbatim text of all OAs, the corresponding applicant rebuttals, the iterative versions of claims at each stage ($\mathcal{C} \rightarrow \mathcal{C}'$), and the complete reference list cited by examiners.

To ensure the high fidelity of PatRe for evaluation, we implement a multi-stage quality control protocol centered on human-expert verification. 
Following an initial automated filtering of documents with excessive noise or metadata errors, then the trained annotators perform a manual audit to verify the structural integrity of prosecution timelines and the logical consistency between cited references and rejection grounds. 
Finally, all personally identifiable information~(such as applicant's name and patent examiner's name) are redacted to adhere to ethic standards, resulting in a high-quality, full-stage benchmark dataset optimized for patent examination modeling and legal reasoning tasks.

\subsection{Dataset Statistics}
\begin{figure*}[!h]
    \centering
    \begin{subfigure}[t]{0.25\textwidth}
        \centering
\includegraphics[width=1\linewidth]{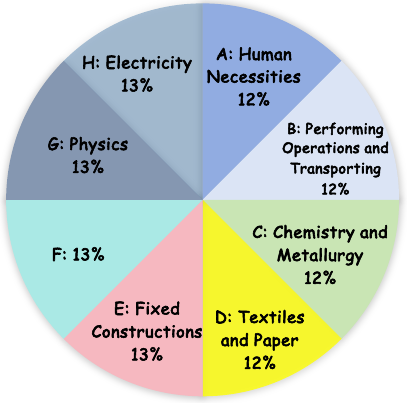}
        \caption{Distribution of IPC sections.}
        \label{figure:ipc}
    \end{subfigure}
    \hfill
    \begin{subfigure}[t]{0.34\textwidth}
        \centering
        \includegraphics[width=1\linewidth]{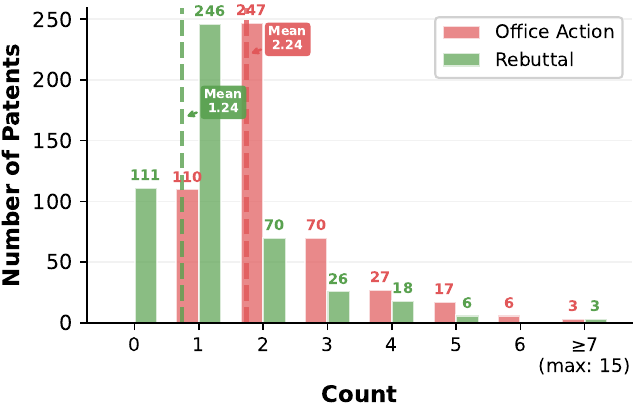}
        \caption{Distribution of Office Action and rebuttal round count.}
        \label{figure:ip_type}
    \end{subfigure}
    \hfill
    \begin{subfigure}[t]{0.35\textwidth}
        \centering
        \includegraphics[width=1\linewidth]{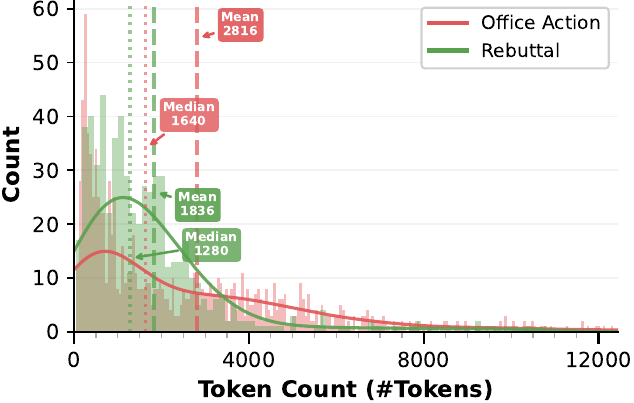}
        \caption{Distribution of Office Action and rebuttal document length.}
        \label{figure:lengh}
    \end{subfigure}
    \caption{Distributions across IPC sections, rounds count, and length. The section F of IPC for Mechanical Engineering, Lighting, Heating, Weapons, and Blasting.}
    \label{figure:data-statics}
\end{figure*}

Our PatRe benchmark comprises the 480 most recent patents, covering all eight sections (A–H) of the International Patent Classification (IPC). 
Each patent includes a complete history of examination records, along with the corresponding OAs, applicant responses, claim revisions, and legal-oriented metadata, such as rejection types and cited reference lists.
As shown in Fig.~\ref{figure:data-statics}, we present (a) the IPC distribution of all patents, (b) the number of rounds of Office Action and rebuttal throughout the full process, and (c) the length distribution of both OA and rebuttal documents.
Given the legal attribution and novelty requirements of the patent examination task, we further provide the distribution of rejection types, OA types, and cited reference counts in Appendix~\ref{appendix:data-statics}.

\section{Experiment}
\subsection{Experimental Setup}

\paragraph{Evaluated Models.}
We benchmark different LLMs covering a broad range of sizes, architectures and families, with model details provided in Appendix~\ref{appendix:models}. These include commercial proprietary models such as GPT series~\citep{Singh2025OpenAIGS}~(GPT-5-mini and GPT-4o-mini), Gemini series~\citep{geminiteam2025geminifamilyhighlycapable}~(Gemini-2.5-Flash) and DeepSeek series~\citep{DeepSeekAI2025DeepSeekV32PT}~(DeepSeek-V3.2). We also include open-source models ranging from 8B to 70B, including LLaMA series~\citep{grattafiori2024llama3herdmodels}, Qwen3.5 series~\citep{qwen35blog} and Gemma3 series~\citep{gemmateam2025gemma3technicalreport} models. 

\paragraph{Implementation Details.}
All proprietary models are assessed via their official APIs, with detailed cost information are in Appendix~\ref{appendix:models}.
We benchmark all open-source models using the vLLM framework~\citep{Kwon2023EfficientMM} on 8 NVIDIA A800 GPUs. 
Given the substantial length of OA and rebuttal, we set the maximum output tokens to each model’s context limit. 
To ensure consistency and reproducibility, we set the temperature to 0.0 across all experiments. 
We use the \texttt{Gemini-3.1-Flash-Lite} as an LLM-as-a-judge evaluator, with the temperature set to 0.0 for consistent evaluation. 
Detailed prompts are in Appendix~\ref{appendix:prompts}. 
We extract additional labels such as rejection type and citations from the generated documents using regular expressions.

\begin{table*}[!t]
\setlength{\tabcolsep}{7pt}
\resizebox{\textwidth}{!}{%
\begin{tabular}{lcccccccccccc}
\toprule
\multirow{2.5}{*}{\textbf{Model}} 
& \multicolumn{4}{c}{\textbf{OA-DP}} 
& \multicolumn{4}{c}{\textbf{OA-RO}} 
&\multicolumn{4}{c}{\textbf{OA-RS}}\\
\cmidrule(lr){2-5} \cmidrule(lr){6-9} \cmidrule(lr){10-13}
& Dec. & Stat. & R-L  & Ovr.
& Dec. & Stat. & R-L  & Ovr.
& Dec. & Stat. & R-L  & Ovr. \\
\midrule
\rowcolor{orange!15}\multicolumn{13}{c}{\textit{\textbf{Proprietary Models}}} \\
\midrule
GPT-5-mini       &\textbf{51.4}  &45.1  &14.8  &\textbf{4.56}  &\textbf{50.0}  &49.2  &17.4  &\textbf{4.89}  &\underline{52.7}  &42.5  &15.4  &\textbf{5.39}  \\
Gemini-2.5-Flash &\underline{50.0}  &\underline{46.6}  &\underline{16.4}  &\underline{4.26}  &46.4  &\textbf{56.3}  &\textbf{20.5}  &\underline{4.36}  &\textbf{52.8}  &\textbf{51.5}  &\textbf{20.5}  &\underline{4.37}  \\
DeepSeek-V3.2    &47.6  &\textbf{47.1}  &\textbf{17.3}  &\underline{4.26}  &\underline{49.7}  &\underline{55.4}  &\underline{19.9}  &4.34  &42.2  &47.0  &\underline{18.7}  &3.50  \\
GPT-4o-mini      &24.4  &44.8  &14.0  &3.75  &43.3  &55.1  &15.7  &3.61  &36.3  &\underline{48.9}  &16.4  &3.59  \\
\midrule
\rowcolor{cyan!15}
\multicolumn{13}{c}{\textit{\textbf{Open-Source Models}}}\\
\midrule
LLaMA3.1-8B-it   &41.4   &42.5  &\textbf{19.3} &3.12  &\underline{45.6} &46.2  &\textbf{24.2}  &3.07  &39.9  &41.0  &\textbf{23.9}  &3.19  \\
Qwen3.5-9B       &41.7  &38.7  &17.7  &\underline{4.07}  &44.3  &48.7  &20.0  &\underline{4.11}  &\underline{43.4}  &43.1  &18.8  &\underline{4.11}  \\
Gemma3-12B-it    & \underline{45.1}   &42.7  &15.3 &3.68 &43.4  &52.6  & 19.6 &3.59&41.6  &42.6  &19.8  &3.51  \\
Gemma3-27B-it    &39.5   &\underline{43.9}  &14.8  &3.75  &45.4  &\textbf{57.7}  &18.0  &3.65 &38.1 &\textbf{49.8} &18.9 &3.68   \\
Qwen3.5-27B      &\textbf{48.8}  &43.7  &\underline{18.3}  &\textbf{4.35}  &\textbf{47.6}  &\underline{56.3}  &\underline{21.4}  &\textbf{4.37}  &\textbf{50.5}  &47.7  &\underline{20.0}  &\textbf{4.30}  \\
LLaMA3.3-70B-it  &10.2  &\textbf{46.3}  &15.1  &3.45  &\text{ }\text{ }9.7  &54.7  &16.6  &3.40  &21.8  &\underline{48.9}  &17.6  &3.50  \\
\bottomrule
\end{tabular}
}
\caption{Results for Office Action generation. Where \textit{Dec.} denotes Decision Accuracy, \textit{Stat.} denotes Statute Precision, \textit{R-L} denotes Rouge-L, and \textit{Ovr.} denotes the average LLM-as-a-judge score.}
\label{tab:oa_main}
\end{table*}

\begin{table*}[t]
\setlength{\tabcolsep}{4pt}
\resizebox{\textwidth}{!}{%
\begin{tabular}{lccccccccccccccc}
\toprule
\multirow{2.5}{*}{\textbf{Model}} 
& \multicolumn{5}{c}{\textbf{OA-DP}} 
& \multicolumn{5}{c}{\textbf{OA-RO}} 
& \multicolumn{5}{c}{\textbf{OA-RS}} \\
\cmidrule(lr){2-6} \cmidrule(lr){7-11} \cmidrule(lr){12-16}
& Sou. & Cla. & Con. & Com. & Sty.
& Sou. & Cla. & Con. & Com. & Sty.
& Sou. & Cla. & Con. & Com. & Sty. \\
\midrule
\multicolumn{16}{c}{\cellcolor{orange!15}\textit{\textbf{Proprietary Models}}} \\
\midrule
GPT-5-mini       &\textbf{2.46}  &\textbf{7.67}  &\textbf{2.16}  &2.28  &\textbf{8.22}  &\textbf{3.07}  &\textbf{7.77}  &\textbf{2.65}  &\textbf{3.14}  &\textbf{7.84}  &\textbf{3.43}  &\textbf{7.89}  &\textbf{3.09}  &\textbf{3.86}  &\textbf{8.69}  \\
Gemini-2.5-Flash &\underline{2.38}  &7.41  &1.67  &\underline{2.30}  &\underline{7.52}  &\underline{2.60}  &7.21  &1.79  &2.60  &\underline{7.60}  &\underline{2.54}  &7.24  &\underline{1.74}  &\underline{2.74}  &\underline{7.61}  \\
DeepSeek-V3.2    &2.26  &\underline{7.48}  &\underline{1.73}  &\textbf{2.33}  &7.48  &2.39  &\underline{7.43}  &\underline{1.86}  &\underline{2.61}  &7.42  &2.18  &\underline{7.39}  &1.66  &2.48  &7.38  \\
GPT-4o-mini      &2.06  &6.81  &1.34  &1.84  &6.74  &2.05  &6.32  &1.45  &2.02  &6.21  &1.86  &6.43  &1.37  &1.94  &6.33  \\
\midrule
\multicolumn{16}{c}{\cellcolor{cyan!15}\textit{\textbf{Open-Source Models}}} \\
\midrule
LLaMA3.1-8B-it   &1.79  &5.44  &1.26  &1.91  &5.22  &1.78  &5.26  &1.24  &1.98  &5.09  &1.80  &5.45  &1.25  &2.10  &5.35  \\
Qwen3.5-9B       &\underline{1.98}  &\underline{7.28}  &\underline{1.65}  &\underline{2.17}  &\underline{7.27}  &\underline{2.26}  &\underline{6.94}  &\underline{1.84}  &\underline{2.58}  &\underline{6.92}  &\underline{2.22} &\underline{6.99} &\underline{1.78}  &\underline{2.60}    &\underline{6.97}  \\
Gemma3-12B-it    &1.87  &6.66  &1.40  &1.86  &6.61  &1.98  &6.34  &1.36  &2.00  &6.27  &1.76  &6.35  &1.27  &1.89  &6.29  \\
Gemma3-27B-it    &1.80  &6.99  &1.35  &1.66  &6.96  &2.01  &6.51  &1.34  &1.95  &6.45  &1.82  &6.76  &1.27  &1.84  &6.72  \\
Qwen3.5-27B      &\textbf{2.31}  &\textbf{7.53}  &\textbf{1.94}  &\textbf{2.45}  &\textbf{7.53}  &\textbf{2.56}  &\textbf{7.19}  &\textbf{2.09}  &\textbf{2.83}  &\textbf{7.18}  &\textbf{2.41}  &\textbf{7.19}  &\textbf{1.94}  &\textbf{2.77}  &\textbf{7.17}  \\
LLaMA3.3-70B-it  &1.95  &6.21  &1.31  &1.75  &6.04  &1.91  &6.08  &1.28  &1.80  &5.95  &1.91  &6.27  &1.27  &1.86  &6.17  \\

\bottomrule
\end{tabular}%
}
\caption{Detailed LLM-as-a-judge results for Office Action generation on a 1-10 scale. \textit{Sou.} (Soundness), \textit{Cla.} (Clarity), \textit{Con.} (Constructiveness), \textit{Com.} (Completeness), and \textit{Sty.} (Language Style).}
\label{tab:oa_judge_detail}
\end{table*}

\begin{table*}[!h]
\setlength{\tabcolsep}{6pt}
\small 
\resizebox{\textwidth}{!}{\begin{tabular}{l c cc ccccc} 
\toprule
\textbf{Model}&\textbf{Cov.}&\textbf{Rouge-L} & \textbf{Soundness} & \textbf{Clarity} & \textbf{Construct.} & \textbf{Complete.} & \textbf{Lang.-Sty.} & \textbf{Average} \\
\midrule
\multicolumn{9}{c}{\cellcolor{orange!15}\textit{\textbf{Proprietary Models}}} \\
\midrule
GPT-5-mini &\textbf{90.5} &14.9 &\textbf{8.71} &\textbf{9.62} &\textbf{8.78} &\textbf{9.12} &\textbf{9.65} &\textbf{9.18}\\
Gemini-2.5-Flash &\underline{81.5} &16.4 &\underline{7.77} &8.80 &\underline{8.06} &\underline{8.12} &8.96 &8.34\\
DeepSeek-V3.2 &79.1 &\underline{17.9} &7.66 &\underline{9.06} &\underline{8.06} &7.94 &\underline{9.15} &\underline{8.37}\\
GPT-4o-mini &27.8 &\textbf{21.1} &2.82 &6.43 &3.44 &3.01 &6.80 &4.50\\
\midrule
\multicolumn{9}{c}{\cellcolor{cyan!15}\textit{\textbf{Open-Source Models}}} \\
\midrule
LLaMA3.1-8B-it &29.5 &18.4 &2.75 &4.63 &3.08 &3.16 &4.94 &3.71\\
Qwen3.5-9B &\underline{63.4} &17.7 &\underline{6.09} &\underline{8.02} &\underline{6.71} &\underline{6.33} &\underline{8.28} &\underline{7.09}\\
Gemma3-12B-it &30.5 &\underline{21.8} &3.12 &6.19 &3.70 &3.21 &6.67 &4.58\\
Gemma3-27B-it &39.9 &19.5 &4.05 &6.96 &4.74 &4.17 &7.42 &5.47\\
Qwen3.5-27B &\textbf{79.2} &17.5 &\textbf{7.61} &\textbf{8.87} &\textbf{8.05} &\textbf{7.93} &\textbf{8.99} &\textbf{8.29}\\
LLaMA3.3-70B-it &30.9 &\textbf{21.9} &3.08 &6.22 &3.65 &3.37 &6.65 &4.59\\
\bottomrule
\end{tabular}}
\caption{Results for {Rebuttal Generation}. Where \textit{Cov.} denotes Point-wise Coverage. The LLM-as-a-judge quality dimensions, scored on a 1–10 scale, include Soundness, Clarity, Constructiveness (Construct.), Completeness (Complete.), and Language Style (Lang.-Sty.).
}
\label{results:rebuttal}
\end{table*}

\subsection{Main Results}
\label{sec：main-results}

\paragraph{\textit{Observation} 1: Proprietary models, especially GPT-5-mini, demonstrate consistently superior performance across both Office Action and Rebuttal generation tasks.} 
As shown in Table~\ref{tab:oa_main}, \texttt{GPT-5-mini} achieves the highest Decision Accuracy in both OA-DP (51.4\%) and OA-RO (50.0\%), while is the second best in the OA-RS setting (52.7\%). 
This performance extends to rebuttal generation (Table~\ref{results:rebuttal}), where it reports a Point-wise Coverage of 90.5\% and an Soundness score of 8.71. 
Notably, the performance disparity between proprietary and open-source models remains relatively narrow in the structured decisional logic required for Office Action tasks. 
However, a more pronounced gap emerges in rebuttal generation tasks, where proprietary models exhibit a distinct advantage in the technical precision and global logical alignment necessary for effective adversarial reasoning.
This suggests that while open-source models are becoming increasingly viable for categorical patentability determinations, a functional bottleneck persists in their ability to handle the complex linguistic and logical demands of applicant-examiner discourse.

\begin{wrapfigure}{r}{0.45\linewidth}
    \centering
\includegraphics[width=1\linewidth]{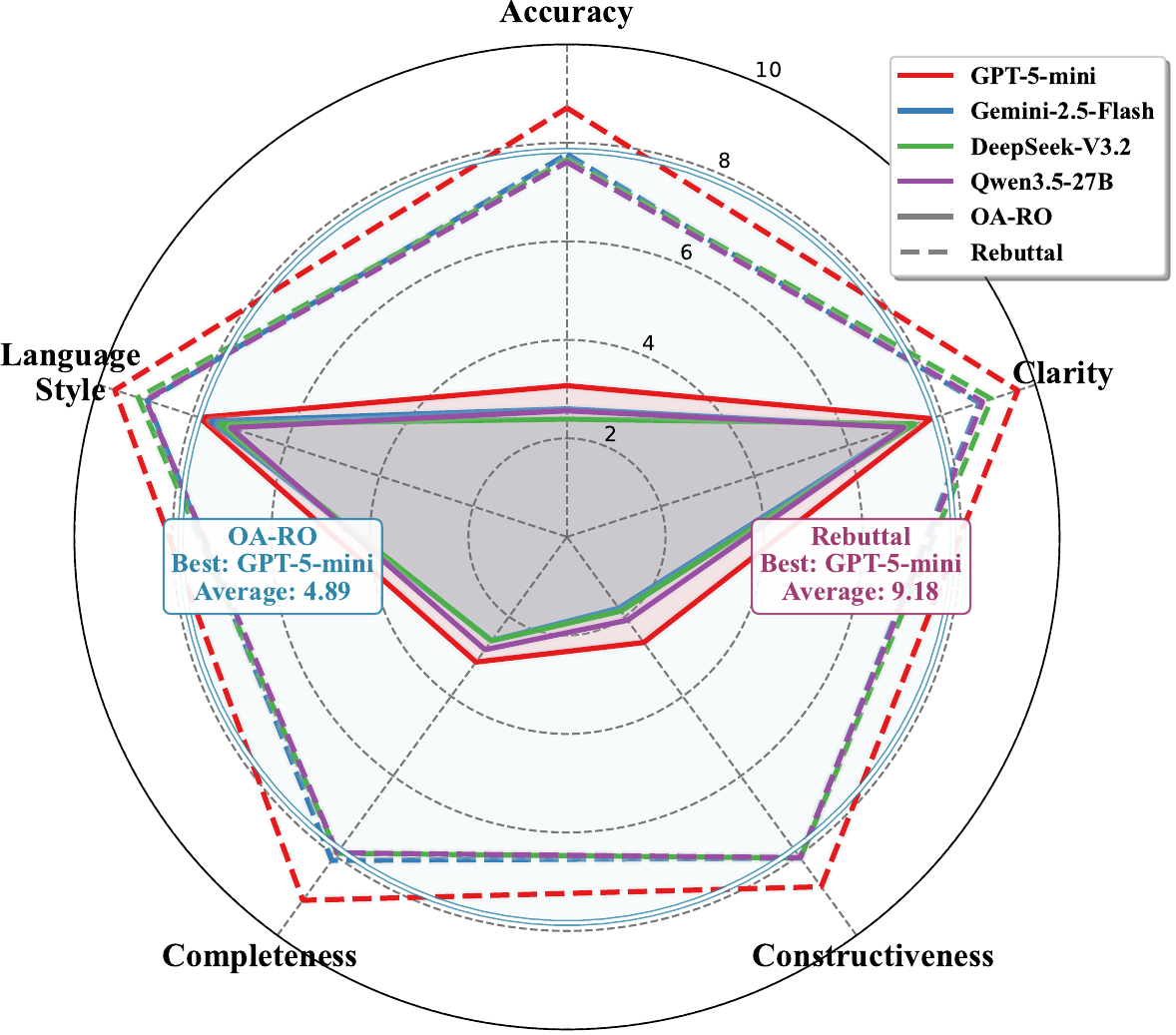}
    \vspace{-1.5em}
    \caption{LLM-as-a-judge scores across 5 dimensions for generated OA-RO and rebuttal.}
    \label{fig:oa-rebuttal-llm-as-a-judge}
    \vspace{-1.2em}
\end{wrapfigure}
\paragraph{\textit{Observation} 2: Models exhibit a significant performance decoupling across various LLM-as-a-Judge dimensions, particularly between surface language style and internal logic.}
We report the detailed LLM-as-a-judge scores across five dimensions in Table~\ref{tab:oa_judge_detail} and Table~\ref{results:rebuttal}. 
These models perform well in language Style and Clarity, but lag far behind in Soundness, Constructiveness, and Completeness.
This confirms a pronounced discrepancy between linguistic form and legal content, where a professional style on surface masks a logic flaw in technical adjudication. 
While this dimensional asymmetry persists across the entire examination lifecycle, its magnitude changes with tasks. 
In Figure~\ref{fig:oa-rebuttal-llm-as-a-judge}, when models transition from proactive examination (OA) to reactive defense (Rebuttal), Soundness and Constructiveness increased more than double, while other dimensions also see marked improvements. 
This suggests that the constraints of legal reasoning are partially mitigated when models respond to explicit grounds. 
But they perform substantially better as responders than as proactive examiners. 
Overall, while lexical professionalism in style is relatively mature, logical reasoning and analysis in Soundness and Completeness remains insufficient in patent examination.

\paragraph{\textit{Observation} 3: The evolution across OA generation settings underscores a performance divergence between statutory citation and substantive adjudication.}
The transition across three OA settings, reveals how information guidance impacts models differently. 
Specifically, while the OA-RO setting acts as an upper bound for Statute Precision (Stat.) due to the availability of oracle references, it does not consistently improve Decision Accuracy (Dec.).
For instance, \texttt{Gemini-2.5-Flash} achieves its peak Stat. in OA-RO, yet its Dec. actually falls below its zero-shot performance in OA-DP.
In OA-RS setting, the top-tier models demonstrate the ability to filter noise and maintain decisional stability.
These observations indicate that, while external evidence strengthens formal legal alignment, it does not inherently reinforce the logical consistency required for accurate patentability determinations.

\subsection{Analysis}

\begin{wrapfigure}{r}{0.52\linewidth}
    \vspace{-1.2em}
    \centering
\includegraphics[width=1\linewidth]{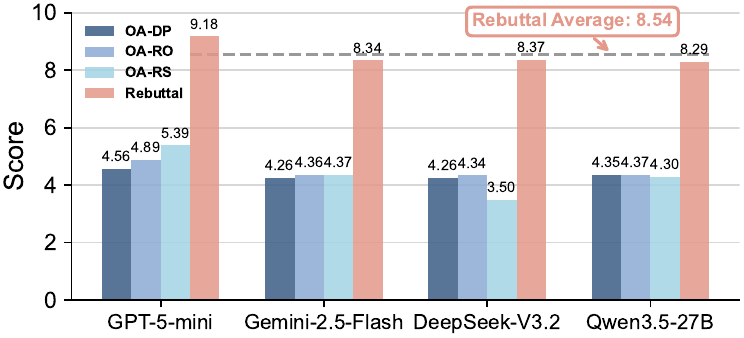}
    \vspace{-1.5em}
    \caption{Office Action vs. Rebuttal generation performance (average LLM-as-a-judge score) of LLMs.}
    \label{fig:oa-vs-rebuttal}
    \vspace{-1em}
\end{wrapfigure}
\paragraph{Finding 1: LLMs exhibit far greater proficiency in \textit{reactive defense} than in \textit{proactive problem discovery}.} 
As illustrated in Figure~\ref{fig:oa-vs-rebuttal}, a significant performance gap exists between the models' roles as applicants and examiners.
Surprisingly, almost models achieve a high \textit{average LLM-as-a-Judge score} during rebuttal generation, but notably lower in OA generation.
For example, while \texttt{DeepSeek-V3.2} and \texttt{Qwen3.5-27B} struggle to exceed an overall score of 4.5 in various OA settings, they consistently achieve scores above 8.0 in rebuttals.
This performance gap indicates a fundamental difference in task complexity: issuing an OA requires proactive problem discovery, where the model must act as an impartial arbitrator to identify statutory defects within dense claims.
Current models naturally aligned with the role of patent attorney, excelling at persuasive advocacy, but struggle with the rigorous, impartial logical deduction required of an examiner.

\begin{wrapfigure}{r}{0.43\linewidth}
    \vspace{-1.2em}
    \centering
\includegraphics[width=1\linewidth,height=0.9\linewidth]{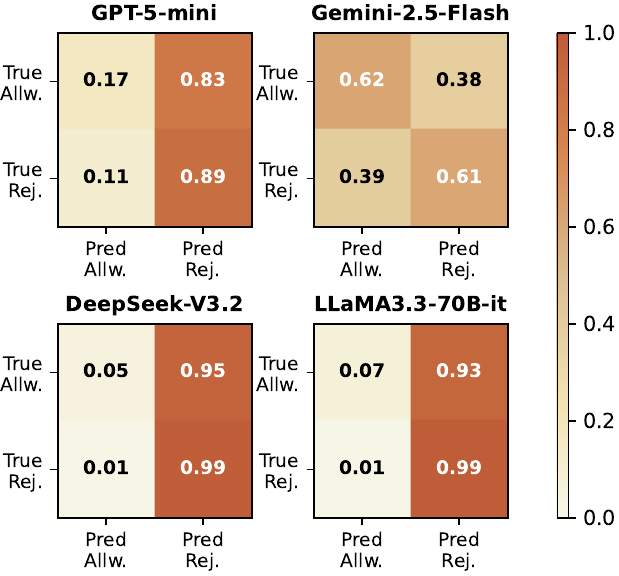}
    \vspace{-1.4em}
    \caption{Heatmap of confusion matrices for allowance (Allw.) vs. rejection (Rej.) classification.}
    \label{fig:confusion_matrix}
    \vspace{-1.2em}
\end{wrapfigure}
\paragraph{Finding 2: A pervasive hyper-critical bias leads to exceptionally high false-rejection rates in Allowance cases.} 
Our quantitative error analysis in Figure~\ref{fig:confusion_matrix}, reveals that when processing applications that should be granted, models frequently hallucinate non-existent conflicts with prior art to justify a rejection. 
Especially for \texttt{LLaMA-3.3-70B-it}, which presents a paradoxical performance profile: while it maintains competitive \textit{Statute Precision}~(54.7\%), its Decision Accuracy suffers a catastrophic deficit, falling as low as 9.7\%.
Specifically, 37.47\% of the errors in \texttt{LLaMA-3.3-70B-it} stem from the premature classification of non-final cases as final rejections. 
This fault-finding behavior reflects an immaturity in legal balance, as models seem to equate rigorous examination with a guaranteed rejection.
This bias is harmful in the patent domain, as models fail to recognize when an invention meets patentability and instead over-scrutinize valid applications, inventing defects to justify rejection.

\begin{wrapfigure}{r}{0.45\linewidth}
    \vspace{-1em}
    \centering
\includegraphics[width=1\linewidth]{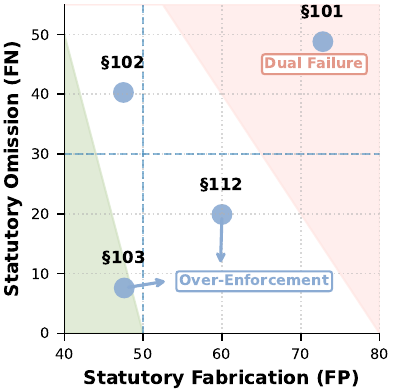}
    \vspace{-1.4em}
    \caption{Statutory error mode landscape: Fabrication vs. Omission.}
    \label{fig:fp+fn}
    \vspace{-1em}
\end{wrapfigure}
\paragraph{Finding 3: LLMs exhibit severe logical inconsistency and over-extension in applying conceptually demanding legal statutes.} 
As illustrated in Figure~\ref{fig:fp+fn},
we categorize statutory errors into two primary modes: \textit{\textbf{statutory fabrication}} (False Positives, FP), where the model predicts a violation that is not present and \textit{\textbf{statutory omission}} (False Negatives, FN), where the model fails to identify a violation that is actually present.
For 35 U.S.C. §101~(Patent Eligibility), models suffer from a \textcolor{dual failure}{\textbf{dual failure}}: they exhibit the highest fabrication rate at 72.8\% while failing to identify 48.8\% of actual eligibility defects.
Similarly, §102~(Novelty) demonstrates significant instability, with high fabrication (FP: 47.5\%) and omission (FN: 40.3\%).
This indicates unstable legal reasoning in LLMs, causing hallucinated oversights.
In contrast, for statutes §103~(Obviousness) and §112~(Definiteness), the error shifts toward aggressive \textcolor{over-enforcement}{\textbf{over-enforcement}}.
While models are less likely to miss a valid rejection (FN: 7.6\% and 19.9\% respectively), they frequently over-extend these rules. Notably, §112 exhibits the second-highest fabrication rate at 60.0\%, while §103 shows a similar over-extension at 47.6\%. 
These findings suggest that while LLMs are prone to over-identifying violations, their ability to accurately distinguish between statutory and non-statutory matter is profoundly insufficient.

\paragraph{Finding 4: Citation accuracy is highly dependent on the quality of external evidence, as models exhibit a strategic reliance on oracle-level retrieval.} 
We evaluate Reference Citation 
\begin{wrapfigure}{r}{0.43\linewidth}
\includegraphics[width=1\linewidth]{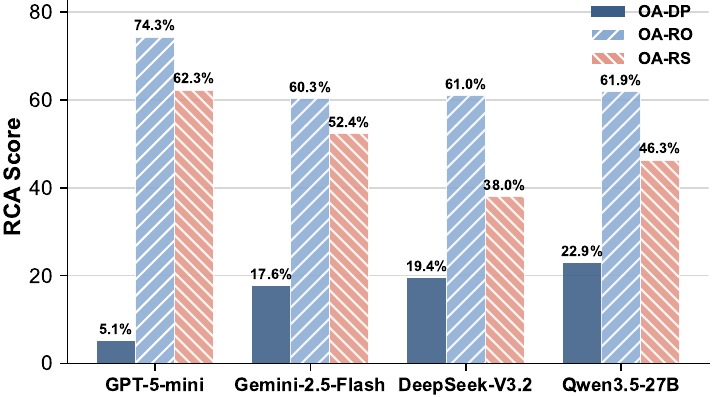}
    \vspace{-1.4em}
    \caption{Reference Citation Accuracy (RCA) performance across OA settings.}
    \label{fig:rca}
    \vspace{-1.4em}
\end{wrapfigure}
Accuracy~(RCA), that captures how accurately the model grounds its technical claims in the provided evidence, \textit{i.e.}, $\text{RCA} = \frac{|R_{\text{pred}} \cap R_{\text{valid}}|}{|R_{\text{pred}}|}$, 
where $R_{\text{pred}}$ denotes the set of references cited in the model's output 
and $R_{\text{valid}}$ represents the valid reference set provided in the context.
As shown in Figure~\ref{fig:rca}, the RCA performance across the three OA settings follows a strictly monotonic trend: $\text{Oracle-Summary} > \text{BM25-Retrieved} \gg \text{No-Ref}$. 
This disparity underscores that statutory citation is a task strictly limited by available reference. 
Therefore, strong reasoning ability cannot replace real external evidence, as models often hallucinate citations when depended only on internal knowledge.

\begin{wraptable}{r}{0.58\textwidth}
\vspace{-10pt}
\setlength{\tabcolsep}{1mm}
\resizebox{\linewidth}{!}{
\begin{tabular}{lccc}
\toprule
\textbf{Metric}
& \textbf{Kendall $\tau$} & \textbf{Pearson $r$} & \textbf{Spearman $\rho$} \\
\midrule
Decision Accuracy   & \textbf{0.4490~(0.0000)} & 0.4863~(0.0000) & 0.5412~(0.0000) \\
Statute Precision   & 0.2852~(0.0040) &  0.2766~(0.0053) & 0.2108~(0.0059) \\
Rouge-L   & 0.1351~(0.0095) & 0.1672~(0.0293)  & 0.1931~(0.0117)\\
{LLM-as-a-judge}   & {0.4480~(0.0000)} & \textbf{0.6808~(0.0000)} & \textbf{0.6231~(0.0000)} \\
\bottomrule
\end{tabular}
}
\caption{Correlation ($\uparrow$) of different metrics with human judgments (\textit{p}-value $\downarrow$).}
\label{tab:metrics-consistency}
\vspace{-15pt}
\end{wraptable}
\paragraph{Finding 5: Lexical metrics exhibit a significant performance decoupling from substantive legal validity.} 
Experimental results in Section~\ref{sec：main-results} show that high textual overlap does not necessarily correlate with decision accuracy or logical consistency.
We observe a trend where models achieving high lexical scores frequently fail to maintain legal correctness~(correlation between Rouge-L and Decision Accuracy: Kendall’s $\tau = 0.0258$).
Conversely, models with lower lexical overlap frequently secure higher judge scores due to superior linguistic precision~(correlation between Rouge-L and LLM-as-a-judge score: Kendall’s $\tau = 0.1440$). 
We conduct \textbf{human evaluations} for the OA-RO and rebuttal generation tasks using three experts, following the same criteria as the LLM-as-a-judge.
Detailed evaluation protocol and results are provided in Appendix~\ref{more-results}.
We compute the alignment of all metrics with human judgments, as shown in Table~\ref{tab:metrics-consistency}. 
The Rouge-L metric remains less consistent with human judgments. 
Human experts show substantial inter-rater agreement, with Pearson’s $r = 0.7285$ and Kendall’s $\tau = 0.5861$. 
We also report inter-rater agreement across different dimensions in Table~\ref{tab:inter_rater_agreement} (See Appendix~\ref{more-results}).
These results demonstrate that traditional n-gram-based metrics are less effective than the LLM-as-a-judge paradigm in capturing the professional nuances of patent examination.

\subsection{Case Study}

To further investigate the performance of LLMs in patent examination, we analyze representative cases of both generated Office Actions and rebuttals. 
Table~\ref{example:case-study-OA-RO}--\ref{example:case-study-Rebuttal} (See Appendix \ref{appendix:cases}) present examples across various stages of the patent examination lifecycle to highlight the gap between professional drafting and model capabilities.
These examples reveal that in OA generation, models often struggle with substantive functional mapping and tend to adopt templated legal discourse to mask a lack of actual technical ambiguity, whereas they demonstrate notable reactive proficiency in Rebuttals by effectively identifying flaws in rejections. 
Overall, these cases demonstrate that current LLMs excel at mimicking the linguistic form of patent examination but struggle with the rigorous, evidence-based reasoning required for precise prior art analysis and statutory evaluation.

\section{Conclusion}
In this work, we introduce PatRe, the first full-stage benchmark for patent Office Action and rebuttal generation for patent examination, which supports the evaluation of LLMs on legal reasoning and technical novelty judgment.
Beyond existing benchmarks that focus on binary classification and static extraction, our PatRe views patent examination as a dynamic, multi-turn process of justification and response. 
It aims to assess the capabilities of frontier LLMs for examination, with the potential for these models to assist patent examiners and applicants in improving efficiency and alleviating the burden of the examination process.
Extensive experiments reveal that current LLMs remain insufficient as independent systems for patent examination and highlight a notable gap between proprietary and open-source models, with open-source models being more suitable for privacy-sensitive settings in this domain.
We envision that PatRe will facilitate future research on patent examination modeling, and plan to extend it to additional jurisdictions and multilingual patents in future work.

\bibliography{ref}

\appendix

\section{More Details about Data Statics}
\label{appendix:data-statics}
We collect patent examination records from the public USPTO data website \footnote{https://data.uspto.gov/home}.
All patents in our dataset were published after 2024, following their Notice of Allowance (NOA) dates.
Given the legal nature of patent examination, we report the distribution of rejection types, including §103 (Obviousness), §112 (Written Description/Enablement), §102 (Lack of Novelty), §101 (Patent-Eligible Subject Matter), and double patenting (DP), as summarized in Table~\ref{tab:rejection_types}. Notably, multiple rejection types may appear within a single case.

\begin{table*}[!h]
\setlength{\tabcolsep}{4mm}
\begin{tabular}{lcc}
\toprule
\textbf{Rejection Type} & \textbf{Count} & \textbf{Percentage} \\
\midrule
103 (Obviousness) & 415 & 40.53\% \\
112 (Written Description/Enablement) & 230 & 22.46\% \\
102 (Lack of Novelty) & 202 & 19.73\% \\
DP (Double Patenting) & 122 & 11.91\% \\
101 (Patent-Eligible Subject Matter) & 55 & 5.37\% \\
\midrule
Total & 1024 & 100.00\% \\
\bottomrule
\end{tabular}
\caption{Distribution of rejection types.}
\label{tab:rejection_types}
\end{table*}

In the Office Action generation task, a complete patent examination history may involve multiple Office Actions. Accordingly, we report the distribution of Office Action types, as summarized in Table~\ref{tab:oa_types}, covering four categories: Notice of Allowance, Non-Final Rejection, Final Rejection, and Ex Parte Quayle Action.

\begin{table*}[!h]
\setlength{\tabcolsep}{4mm}
\begin{tabular}{lcc}
\toprule
\textbf{OA Type} & \textbf{Count} & \textbf{Percentage (\%)} \\
\midrule
Notice of Allowance & 479 & 44.56 \\
Non-Final Rejection & 435 & 40.47 \\
Final Rejection & 152 & 14.14 \\
Ex Parte Quayle Action & 9 & 0.84 \\
\midrule
Total & 1075 & 100.00 \\
\bottomrule
\end{tabular}
\caption{Distribution of Office Action (OA) types.}
\label{tab:oa_types}
\end{table*}

Beyond legal labels, to reflect the novelty requirements in patent examination, we analyze the distribution of cited references within each patent record, as shown in Table~\ref{tab:citation_discussion}.

\begin{table*}[!h]
\setlength{\tabcolsep}{4mm}
\begin{tabular}{lcccc}
\toprule
\textbf{Category} & \textbf{Avg Cited} & \textbf{Avg Discussed} & \textbf{Difference} & \textbf{Ratio (\%)} \\
\midrule
A & 11.33 & 9.60 & 1.73 & 84.71\% \\
B & 8.82 & 6.15 & 2.67 & 69.75\% \\
C & 2.88 & 5.77 & -2.88 & 200.00\% \\
D & 9.32 & 4.32 & 5.00 & 46.33\% \\
E & 9.75 & 4.72 & 5.03 & 48.38\% \\
F & 8.92 & 3.72 & 5.20 & 41.68\% \\
G & 9.70 & 9.52 & 0.18 & 98.11\% \\
H & 8.08 & 6.57 & 1.52 & 81.24\% \\
\midrule
All & 8.60 & 6.30 & 2.30 & 73.78\% \\
\bottomrule
\end{tabular}
\caption{Citation vs. discussion statistics across categories. 
“Avg Cited” and “Avg Discussed” denote the average numbers of cited and explicitly discussed references per patent record, respectively. 
“Difference” is computed as cited minus discussed, and “Ratio” represents the proportion of discussed references relative to cited ones.}
\label{tab:citation_discussion}
\end{table*}

\section{More Details about Evaluated Models}
\label{appendix:models}
We present the details of the evaluated models in Table~\ref{tab:models}, including their sizes, maximum context lengths, maximum output lengths, and access methods.
For proprietary models and DeepSeek-V3.2 (which we treat as proprietary due to computing resource constraints), we evaluate them via their official APIs to ensure a fair and consistent comparison. 
We further report the total costs incurred by these proprietary models on the PatRe benchmark, as summarized in Table~\ref{tab:costs}.
For open-source models with sizes less than or equal to 70B, we deploy them using the vLLM framework~\citep{Kwon2023EfficientMM} on 8 NVIDIA A800 GPUs with the same hyperparameters to ensure a fair comparison.

\begin{table*}[!h]
\setlength{\tabcolsep}{5mm}
\resizebox{\linewidth}{!}{
\begin{tabular}{lccccc}
\toprule
\textbf{Model} & \textbf{Size} & \textbf{Max Context} & \textbf{Max Output} & \textbf{Access} \\
\midrule
GPT-5-mini~\citep{Singh2025OpenAIGS} & -- & 400K & 128K & OpenAI API \\   
GPT-4o-mini~\citep{hurst2024gpt} & -- & 128K & 16K & OpenAI API \\
Gemini-2.5-Flash~\citep{geminiteam2025geminifamilyhighlycapable} & -- & 1M & 64K & Google API \\
DeepSeek-V3.2~\citep{DeepSeekAI2025DeepSeekV32PT} & 671B & 128K & 64K & DeepSeek API \\
Llama3.1-8B-Instruct~\citep{grattafiori2024llama3herdmodels} & 8B & 128K & -- & Weights \\  
Qwen3.5~\citep{qwen35blog} & 9/27B & 1M & 64K & Weights \\ 
Gemma3-Instruct~\citep{gemmateam2025gemma3technicalreport} & 9/27B & 128K & -- & Weights \\
Llama3.3-70B-Instruct~\citep{grattafiori2024llama3herdmodels} & 70B & 128K & -- & Weights \\  
\bottomrule
\end{tabular}
}
\caption{The overview of evaluated models.}
\label{tab:models}
\end{table*}

\begin{table*}[!h]
\begin{tabular}{lcccc}
\toprule
\textbf{Model} & GPT-5-mini & GPT-4o-mini & Gemini-2.5-Flash & DeepSeek-V3.2  \\
\midrule
\textbf{Cost~(USD)} & 37.85  & 11.08  & 14.13  & 23.49 \\
\bottomrule
\end{tabular}
\caption{Cost of different models (accessed via API).}
\label{tab:costs}
\end{table*}

\section{More Results}
\label{more-results}
\paragraph{Human Evaluation Setup.}
We employ three PhD students specializing in Intellectual Property (IP), who are trained and have academic backgrounds in IP, equipping them with foundational knowledge of both the technical and legal aspects of IP. 
All human evaluators are provided with the same evaluation criteria as those in Figure~\ref{prompt:llm-oa-generation} and Figure~\ref{prompt:llm-rebuttal-generation}.
We sample 100 generated OAs and 100 generated rebuttals across 7 models and all IPC sections, and evaluate them under a blind setting, where the evaluators are unaware of which model generated each instance.

\begin{table*}[!h]
\begin{tabular}{lccc}
\toprule
\textbf{Dimension} & \textbf{Pearson} $r$ & \textbf{Spearman} $\rho$ & \textbf{Kendall} $\tau$ \\
\midrule
Accuracy & 0.6061 & 0.6545 & 0.5380 \\
Clarity & 0.6899 & 0.6863 & 0.5718 \\
Completeness & 0.6858 & 0.7259 & 0.5819 \\
Constructiveness & 0.5929 & 0.6621 & 0.5486 \\
Language Style & 0.4676 & 0.4609 & 0.3787 \\
\midrule
Average Score & 0.7285 & 0.7715 & 0.5861 \\
\bottomrule
\end{tabular}
\caption{Inter-rater agreement between three human experts across five evaluation dimensions, measured using Pearson’s $r$, Spearman’s $\rho$, and Kendall’s $\tau$.}
\label{tab:inter_rater_agreement}
\end{table*}

\paragraph{Detailed Human Evaluation Result.}
We provide detailed human evaluation results across five dimensions for generated Office Actions and rebuttals, as shown in Table~\ref{results:oa_rebuttal}. 
Notably, we provide detailed inter-rater agreement among three human experts across five dimensions, as shown in Table~\ref{tab:inter_rater_agreement}, demonstrating strong consistency across these dimensions.

\begin{table*}[!h]
\setlength{\tabcolsep}{5pt}
\small
\resizebox{\textwidth}{!}{
\begin{tabular}{lcccccccccccc}
\toprule
\multirow{2.5}{*}{\textbf{Model}}
& \multicolumn{6}{c}{\textbf{OA-RO Generation}} 
& \multicolumn{6}{c}{\textbf{Rebuttal Generation}} \\

\cmidrule(lr){2-7} \cmidrule(lr){8-13}

& {Sound.} & {Clar.} & {Const.} & {Comp.} & {Lang.} & {Avg.}
& {Sound.} & {Clar.} & {Const.} & {Comp.} & {Lang.} & {Avg.} \\

\midrule
\multicolumn{13}{c}{\cellcolor{orange!15}\textit{\textbf{Proprietary Models}}} \\
\midrule

GPT-5-mini 
&\underline{2.42}  &4.50  &2.69  &2.69  &\underline{5.65}  &\underline{3.59}  
&\textbf{5.94}  &\underline{6.75}  &\textbf{6.69} &\textbf{7.06}  &\textbf{7.81}  &\textbf{6.85} \\

Gemini-2.5-Flash 
&2.36  &\underline{4.86}  &\underline{2.78}  &\underline{3.11}  &5.31  &3.68  
&\underline{5.75}  &\textbf{7.21}  &\underline{6.04}  &\underline{6.92} &\underline{7.42}  &\underline{6.67} \\

DeepSeek-V3.2 
&\textbf{2.71}  &\textbf{5.71}  &\textbf{3.32}  &\textbf{3.37}  &\textbf{6.55}  &\textbf{4.33}  
& 4.00 &6.58 & 4.00 & 4.75 & 5.92 &5.05 \\

GPT-4o-mini 
&1.58  &3.96  &1.85  &1.96  &5.46  &2.97  
&3.38  &6.31  &3.38  &4.31  &6.63  &4.80 \\

\midrule
\multicolumn{13}{c}{\cellcolor{cyan!15}\textit{\textbf{Open-Source Models}}} \\
\midrule

LLaMA3.1-8B-it 
&1.55  &3.77  &1.45  &1.73  &4.55  & 2.61 
&3.92  &5.67  &\underline{4.17}  &\underline{5.33}  &6.58  &5.13 \\

Gemma3-12B-it 
&\textbf{1.79}  &\textbf{4.68}  &\underline{1.79}  &\underline{2.07}  &\underline{5.50}  &\textbf{3.16}  
&\underline{4.12}  &\underline{5.92}  &4.08  &4.85  &\underline{6.71}  &\underline{5.16}\\

Gemma3-27B-it 
&\underline{1.71}  &\underline{4.21}  &\textbf{1.96}  &\textbf{2.11}  &\textbf{5.71}  &\underline{3.14}  
&\textbf{4.61}  &\textbf{6.96}  &\textbf{4.71}  &\textbf{6.18}  &\textbf{7.36}  &\textbf{5.96}  \\
\bottomrule
\end{tabular}
}
\caption{Human evaluation results for OA-RO and Rebuttal generation. Scores are on a 1--10 scale across five dimensions: Soundness~(Sound.), Clarity~(Clar.), Constructiveness~(Const.), Completeness~(Comp.), and Language Style~(Lang.).}
\label{results:oa_rebuttal}
\end{table*}

\section{Detailed Case Study}
\label{appendix:cases}

To further investigate the LLMs performance in patent examination, we analyze specific cases across different task settings, we provide detailed generated Office Action cases, including OA-RO (Table~\ref{example:case-study-OA-RO}), OA-RS (Table~\ref{example:case-study-OA-RS}) and OA-DP (Table~\ref{example:case-study-OA-DP}) settings, as well as a generated rebuttal case (Table~\ref{example:case-study-Rebuttal}).

Across different settings, we observe consistent limitations in model performance on patent examination tasks. 
In the OA-RO setting, even with access to oracle references, models can identify relevant prior art but often fail to perform rigorous claim-to-art mapping, instead reducing complex technical distinctions to superficial similarities.
In the OA-RS setting, although the correct reference may be present, models tend to incorporate irrelevant documents into combinations, leading to unsupported mappings and unstable novelty judgments. 
In the OA-DP setting, when external evidence is unavailable, models frequently produce incorrect rejections that are not grounded in the claim text or prior art. 
In contrast, in rebuttal generation, models show stronger performance by effectively responding to given rejections and constructing coherent counter-arguments. 
Overall, these results suggest that while external evidence improves citation accuracy, the core reasoning required for precise claim-to-prior-art analysis remains a major bottleneck, and current models perform better in reactive argumentation than in proactive examination.

\begin{table*}[!t]
\scriptsize
\sethlcolor{orange!20} 
\scalebox{1}{\begin{tabular}{p{\linewidth}}
\toprule
\textbf{Task}: \textit{Reference-Oracle Generation}~(OA-RO) \\
\midrule
\hl{\textbf{[Initial]}}~\textbf{Input:}
\begin{itemize}
    \item \textbf{CLAIMS:} \newline
    WHAT IS CLAIMED IS: 
    1. A system for just-in-time (JIT) game development and gameplay, the system comprising:a game engine: an oracle comprising one or more Al models~[ ... Claims 2-30 omitted for brevity ... ]
    \item \textbf{Oracle Art:} \newline
   \textbf{ Ref US20190362589:} 1. A system for coincident gambling hybrid gaming having coincident gambling events[...]
    \newline\textbf{Ref US20240390801A1:} A game server for providing updates related to an online game
    [ ... Other pertinent references omitted ... ]
    \item \textbf{Task summary:} \newline Act as a USPTO examiner. Given current claims and an oracle reference set, evaluate statutory rejections including but not limited to 35 U.S.C. 101/102/103/112 and Double Patenting (DP). Select pertinent references, perform element-by-element mapping, and justify reference selection. Adhere to MPEP standards and avoid disclaimers.
\end{itemize}
\\
\hl{\textbf{[Initial]}}~\textbf{Output:}\\
\begin{minipage}{\linewidth}
\begin{list}{}{
        \setlength{\leftmargin}{2em} 
        \setlength{\parsep}{0pt}
        \setlength{\topsep}{0pt}
    }
\item %
DETAILED ACTION\\
OPENING STATEMENT\\
The present application is examined for compliance with 35 U.S.C. § 101, 102, 103, and 112. It is determined that claims 1-30 are not allowable in their present form.\newline
CLAIM REJECTIONS - 35 U.S.C. 103\newline
Claims 1-30 are rejected under 35 U.S.C. § 103 as being unpatentable over \textcolor{green!60!black}{[Ref US20190362589]} in combination with \textcolor{green!60!black}{[Ref US20240390801A1]}.\newline
Ref US20190362589 discloses a system for coincident gambling hybrid gaming having coincident gambling events and game events provided on a computing device. 
Ref US20240390801A1 discloses a game server for providing updates related to an online game, wherein the game server is configured to sandbox potentially malicious event-driven code.[...]\newline
Specifically, the following limitations are met: [ ... Detailed element-by-element mapping for Claims 1-30 omitted for brevity ... ]\newline
CONCLUSION: Claims 1-30 are rejected as being unpatentable under 35 U.S.C. § 103 over Ref US20190362589 in combination with Ref US20240390801A1. The applicant could  respond to these rejections with arguments or amendments that overcome the references.
\end{list}
\end{minipage} \\
\midrule
\hl{\textbf{[Intermediate]}}~\textbf{Input:}
\begin{itemize}
    \item \textbf{CLAIMS:} \newline
    1. (Currently Amended) A system for just-in-time (JIT) game development and gameplay, the system comprising:a game engine; oracle comprising one or more Al models [ ... Claims 2-30 omitted for brevity ... ]
    \item \textbf{Oracle Art:} \newline Ref US20240390801A1: A game server for providing updates related to an online game, wherein the game server is configured to sandbox potentially malicious event-driven code [ ... Other pertinent references omitted for brevity ... ]
    \item \textbf{rebuttal:} \newline REMARKS In response to the non-final Office Action dated March 26, 2025, Applicant amends claims 1-7, 11, 19-27, 29, and 30 [...]
    \item \textbf{Examination History:} \newline Previous Round Office Action: The present application, filed on or after March 16, 2013, is being examined under the first inventor to file provisions of the AIA [...]
    \item \textbf{Task summary:} \newline Act as a USPTO examiner to conduct a formal re-examination by evaluating amended claims and applicant rebuttals against the established prosecution history. The process requires a critical selection of prior art to address specific traversals, applying Strict Statute Gating to determine whether to maintain or withdraw previous rejections across all applicable statutory grounds. Each argument must be assessed as Persuasive or Unpersuasive with technical reasoning grounded in a rigid Scope Lock, ensuring the final report acknowledges claim amendments, identifies any new matter violations, and presents updated rejections within a realistic, structured USPTO-style action.
\end{itemize}\\
\hl{\textbf{[Intermediate]}}~\textbf{Output:}\\
\begin{minipage}{\linewidth}
\begin{list}{}{
        \setlength{\leftmargin}{1.5em} 
        \setlength{\parsep}{0pt}
        \setlength{\topsep}{0pt}
    }
\item %
DETAILED ACTION\newline
The application is currently pending. Amendments have been made to claims 1-7, 11, 19-27, 29, and 30. Claims 1 and 26 are in independent form.
\newline
RESPONSE TO ARGUMENTS
\newline
The applicant’s arguments regarding 35 U.S.C. § 112 rejections for claims 4 and 7 are persuasive. [...] The applicant’s arguments regarding 35 U.S.C. §§ 102 and 103 rejections are unpersuasive. While the applicant attempts to distinguish the claimed invention from Ignatchenko, the asserted reference, in combination with Arnone, renders the claims unpatentable. The applicant’s attempts to characterize the claimed features as “notorious well-known” do not negate the teachings of the cited references.
\newline
WITHDRAWN REJECTIONS\newline
Claims 4 and 7 – 35 U.S.C. 112\newline
CLAIM REJECTIONS - 35 U.S.C. 102\newline
Claims 1-10, 23-28, and 30 are rejected under 35 U.S.C. 102(a)(1) as being anticipated by Ignatchenko (US 20240390801 A1).  [...] The claimed features are all disclosed within Ignatchenko.\newline
CLAIM REJECTIONS - 35 U.S.C. 103\newline
Claims 11-13, 21, 22, and 29 are rejected under 35 U.S.C. 103 as being unpatentable over Ignatchenko (US 20240390801 A1) in combination with Arnone (US 20190362589 A1). [...] It would have been obvious to one of ordinary skill in the art to combine these references to achieve the claimed invention.\newline
[ ... Rejections for Claims 14-20 and formal Sections 112, Double Patenting omitted for brevity ... ]\newline
CONCLUSION\newline
Claims 1-10, 11-13, 14-20, 21, 22, 23-28, 29, and 30 are currently rejected under 35 U.S.C. 102 and 103. Claims 1 and 26 are in condition for allowance, pending the applicant’s response to the rejections.
\end{list}
\end{minipage} \\
\bottomrule
\end{tabular}}
\caption{Example of generated Office Action under Reference-Oracle Generation~(OA-RO) setting.}
\label{example:case-study-OA-RO}
\end{table*}

\begin{table*}[!t]
\scriptsize
\sethlcolor{orange!20} 
\scalebox{1}{\begin{tabular}{p{\linewidth}}
\toprule
\textbf{Task}: \textit{Retrieval-Simulated Generation}~(OA-RS) \\
\midrule
\hl{\textbf{[Initial]}}~\textbf{Input:}
\begin{itemize}
    \item \textbf{CLAIMS:} \newline 1. A system for at least partially filling an orchestra pit or extending a stage, comprising [ ... Claims 2-12 omitted for brevity ... ] 
    \item \textbf{Candidate Pool:} \par-- \textbf{Oracle Art:} \textbf{Ref US4332116}: 1. A building structure formed from prefabricated building components, the plan view of which is a polygon havinga central support column [ ... Other pertinent references omitted for brevity ... ]\par
-- \textbf{BM25 Set:} \textbf{Ref US8117787B2}: A length-adjustable construction support, comprising[...] [ ... Other pertinent references omitted for brevity ... ]\par \item \textbf{Task summary:}\newline Act as a USPTO examiner. Given claims and a noisy pool of 10 candidates (BM25 + Oracle), evaluate patentability across various statutory grounds (e.g., 102, 103, 112). Distinguish pertinent prior art from irrelevant noise. Ground all technical reasoning in the provided inputs and generate a formal Office Action report.
\end{itemize}
\\
\hl{\textbf{[Initial]}}~\textbf{Output:}\\
\begin{minipage}{\linewidth}
\begin{list}{}{
        \setlength{\leftmargin}{2em} 
        \setlength{\parsep}{0pt}
        \setlength{\topsep}{0pt}
    }
\item %
DETAILED ACTION\newline
OPENING STATEMENT\newline
This is an examination of claims 1-20 of your application. A review of the claims and prior art has revealed certain objections. The following analysis details these objections and provides a basis for rejection.
\newline
CLAIM REJECTIONS - 35 U.S.C. 103\newline
Claims 1-20 are rejected under 35 U.S.C. § 103 as unpatentable over \textcolor{green!60!black}{[US4332116]} in combination with \textcolor{red!70!black}{[US8117787B2]}. US4332116 discloses a building structure formed from prefabricated building components, including a frame assembly with a leg.[...] US8117787B2 discloses a length-adjustable construction support comprising a first pipe, a second pipe, and an inner stopper coupled to the outer circumference of the second pipe. [...]\newline
[ ... Detailed element-by-element mapping for Claims 1-20 omitted for brevity ... ]
\newline
CONCLUSION\newline
Claims 1-20 are rejected as unpatentable under 35 U.S.C. § 103 and 35 U.S.C. § 112(a). Applicant is invited to amend the claims to overcome these rejections. A reasoned explanation of any amendments should be provided.
\end{list}
\end{minipage} \\
\midrule
\hl{\textbf{[Intermediate]}}~\textbf{Input:}
\begin{itemize}
    \item \textbf{CLAIMS:} \newline 1. This listing of claims will replace all prior versions, and listings, of claims in the application [...] 
    \item \textbf{Candidate Pool:} \par-- \textbf{Oracle Art:} \textbf{Ref US11117934}: (A myxoma virus (MYXV) having enhanced anti-cancer activity, wherein the MYXV is genetically engineered to attenuate an activity or expression level of its M153 proteinn[...] [ ... Other pertinent references omitted for brevity ... ]\par
-- \textbf{BM25 Set:}\textbf{ Ref US20180042847A1}: A method of treating and/or reducing the incidence of chronic obstructive pulmonary disease (COPD)n[...] [ ... Other pertinent references omitted for brevity ... ]
    \item \textbf{rebuttal:}REMARKS Claims 1-20 are currently pending and under consideration[...]
    \item \textbf{Examination History:}revious Round Office Action: DETAILED ACTION Notice of Pre-AIA  or AIA  Status The present application[...]
\par 
\textbf{Task summary:} Act as a USPTO examiner to conduct a formal re-examination by navigating a high-entropy candidate pool containing both BM25-retrieved noise and Oracle-level references. The process requires the model to autonomously evaluate and select the most pertinent prior art to address amended claims and applicant rebuttals, applying Active Statute Gating to determine whether to maintain or withdraw previous rejections across all applicable statutory grounds. Each argument must be assessed as Persuasive or Unpersuasive through evidence-first mapping and a strict Scope Lock on provided technical summaries, ensuring the final report acknowledges amendments, identifies potential new matter violations, and presents updated rejections within a structured, realistic USPTO-style Office Action.
\end{itemize}\\
\hl{\textbf{[Intermediate]}}~\textbf{Output:}\\
\begin{minipage}{\linewidth}
\begin{list}{}{
        \setlength{\leftmargin}{2em} 
        \setlength{\parsep}{0pt}
        \setlength{\topsep}{0pt}
    }
\item %
DETAILED ACTION\newline
The present application is being examined under the first inventor to file provisions of the AIA. 
\newline
RESPONSE TO ARGUMENTS\newline
The applicant argues that the double patenting rejection has been overcome by filing a Terminal Disclaimer. This argument is Persuasive as the Terminal Disclaimer is properly filed and the reference patent, US11117934, is commonly owned.\newline
No rejection under 35 U.S.C. 102 is made at this time. However, a rejection under 35 U.S.C. 103 is made as the claims are obvious over \textcolor{green!60!black}{[Ref US11117934]}. The reference teaches a MYXV with a genetic modification at its M153 locus [...] similar to the claims of the present application.
[ ... Detailed reference selection and evidence-first mapping omitted for brevity ... ]
The nonstatutory double patenting rejection is based on a judicially created doctrine grounded in public policy so as to prevent the unjustified or improper timewise extension of the “right to exclude” granted by a patent and to prevent possible harassment by multiple assignees...
\newline
CONCLUSION
\newline
The nonstatutory double patenting rejection is withdrawn, and the application is still subject to examination under 35 U.S.C. 102 and 103. A rejection under 35 U.S.C. 103 is made.
\end{list}
\end{minipage} \\
\bottomrule
\end{tabular}}
\caption{Example of generated Office Action under Retrieval-Simulated Generation~(OA-RS) setting.}
\label{example:case-study-OA-RS}
\end{table*}

\begin{table*}[!t]
\scriptsize
\sethlcolor{orange!20} 
\scalebox{1}{\begin{tabular}{p{\linewidth}}
\toprule
\textbf{Task}: \textit{Direct Prompting Generation}~(OA-DP) \\
\midrule
\hl{\textbf{[Initial]}}~\textbf{Input:}
\begin{itemize}
    \item \textbf{CLAIMS}: \newline 1: An extendable support leg comprising:a top leg tube (1), a middle leg tube (2) and an end leg tube (3) arranged axially telescopically and sequentially from outside to inside [...] 
    \item \textbf{Task summary:} Act as a USPTO examiner. Given current claims without any prior art documents, evaluate statutory rejections (e.g., 101, 102, 103, 112, or DP). Rely solely on your internal technical knowledge and common sense available to a PHOSITA. State technical reasoning before deciding on allowance or rejection.
\end{itemize}
\\
\hl{\textbf{[Initial]}}~\textbf{Output:}\\
\begin{minipage}{\linewidth}
\begin{list}{}{
        \setlength{\leftmargin}{2em} 
        \setlength{\parsep}{0pt}
        \setlength{\topsep}{0pt}
    }
\item %
DETAILED ACTION\newline
OPENING STATEMENT\newline
The present examination has been completed with respect to claims 1-10. The claims are examined for compliance with 35 U.S.C. § 112(a) and (b).
\newline
\textcolor{red!70!black}{[CLAIM REJECTIONS - 35 U.S.C. 112(b)]}
\newline
Claims 1-10 are rejected under 35 U.S.C. § 112(b) as being indefinite. The specification does not provide sufficient support to enable a person having ordinary skill in the art to make and use the claimed invention. Specifically, the description of the first clamping assembly in claim 10 lacks clarity and precision. The relationship between the various components (drive tube, upper locking piece, locking body, and expansion member) and their function in achieving clamping is not adequately explained. The description of the expansion member's behavior, transitioning between a tightened and loosen state based on the movement of the upper locking piece, is vague and lacks sufficient detail to allow for practical implementation. The specification does not provide a clear understanding of how the expansion member interacts with the inner wall of the top leg tube, nor does it provide sufficient detail regarding the forces and mechanisms involved...
\newline
CONCLUSION\newline
Claims 1-10 are currently rejected under 35 U.S.C. § 112(b) for indefiniteness. The applicant is invited to amend the claims to provide a clear and concise description of the invention, enabling a person having ordinary skill in the art to make and use the claimed invention without undue experimentation.
\end{list}
\end{minipage} \\
\midrule
\hl{\textbf{[Intermediate]}}~\textbf{Input:}
\begin{itemize}
    \item \textbf{Claim 1, 2, 3, ...} (AMENDMENTS TO THE CLAIMS: The following listing of claims will replace all prior versions and listings of claims in this application ...). 
\item \textbf{rebuttal:}REMARKS: Favorable reconsideration of this application is respectfully requested in view of the foregoing amendments and the following remarks...
\item \textbf{Examination History:}Previous Round Office Action: DETAILED ACTION Notice of Pre-AIA  or AIA  Status The present application, filed on or after March 16, 2013...
\item \textbf{Task summary:} Act as a USPTO examiner to conduct a formal re-examination by evaluating amended claims and applicant rebuttals relying solely on internal technical knowledge and PHOSITA common sense in the absence of external prior art. The process requires applying Active Statute Gating to maintain or withdraw existing rejections across all applicable statutory grounds, assessing each argument as Persuasive or Unpersuasive through an Evidence-First Logic derived from technical principles. Adhering to a strict Scope Lock, the model must acknowledge amendments, identify any new matter violations, and present updated rejections—specifically focusing on inherent obviousness or indefiniteness—within a structured, realistic USPTO-style Office Action.
\end{itemize}\\
\hl{\textbf{[Intermediate]}}~\textbf{Output:}\\
\begin{minipage}{\linewidth}
\begin{list}{}{
        \setlength{\leftmargin}{2em} 
        \setlength{\parsep}{0pt}
        \setlength{\topsep}{0pt}
    }
\item %
DETAILED ACTION\newline
This Office Action is in response to the Amendment and Remarks filed on [Date of Submission]. Claims 13-29 are pending. Claims 13 and 28 are currently amended. Claims 1-12 are canceled.
\newline
The present examination has been completed with respect to claims 1-10. The claims are examined for compliance with 35 U.S.C. § 112(a) and (b).
\newline
\textcolor{red!70!black}{[CLAIM REJECTIONS - 35 U.S.C. 112(b)]}
\newline
Claims 1-10 are rejected under 35 U.S.C. § 112(b) as being indefinite. The specification does not provide sufficient support to enable a person having ordinary skill in the art to make and use the claimed invention. Specifically, the description of the first clamping assembly in claim 10 lacks clarity and precision. The relationship between the various components (drive tube, upper locking piece, locking body, and expansion member) and their function in achieving clamping is not adequately explained. The description of the expansion member's behavior, transitioning between a tightened and loosen state based on the movement of the upper locking piece, is vague and lacks sufficient detail to allow for practical implementation. The specification does not provide a clear understanding of how the expansion member interacts with the inner wall of the top leg tube, nor does it provide sufficient detail regarding the forces and mechanisms involved...
\newline
CONCLUSION\newline
Claims 1-10 are currently rejected under 35 U.S.C. § 112(b) for indefiniteness. The applicant is invited to amend the claims to provide a clear and concise description of the invention, enabling a person having ordinary skill in the art to make and use the claimed invention without undue experimentation.
\end{list}
\end{minipage} \\
\bottomrule
\end{tabular}}
\caption{Example of generated Office Action under Direct-Prompting Generation~(OA-RO) setting.}
\label{example:case-study-OA-DP}
\end{table*}

\begin{table*}[!t]
\scriptsize
\sethlcolor{orange!20} 
\scalebox{1}{\begin{tabular}{p{\linewidth}}
\toprule
\textbf{Task}: \textit{Rebuttal Generation}\\
\midrule
\hl{\textbf{Input:}}
\begin{itemize}
    \item \textbf{Claim 1, 2, 3, ...} (1. A pharmaceutical composition comprising ...). 
    \item \textbf{Oracle Art (Cited in OA):} Ref US11117934 (A pharmaceutical composition comprising:) a myxoma virus (MYXV) comprising a genetic modification at its M153 locus and a transgene encoding a cytokine or a cell matrix protein...) and ...
    \item
\textbf{Reference OA:}DETAILED ACTION Notice of Pre-AIA  or AIA  Status The present application, filed on or after March 16, 2013, is being examined under the first inventor to file provisions of the AIA...
\end{itemize}
\\
\hl{\textbf{Output:}}\\
\begin{minipage}{\linewidth}
\begin{list}{}{
        \setlength{\leftmargin}{2em} 
        \setlength{\parsep}{0pt}
        \setlength{\topsep}{0pt}
    }
\item %
RESPONSE TO OFFICE ACTION
\newline
AMENDMENT AND REMARKS
\newline
I. Status of the Claims...\newline
IV. Section 102 Rejections\newline
The Examiner rejects claims 1-20 over claims 1-4, 8, 16-23, 26, 27 of U.S. Patent No. 11,117,934 (the “’934 Patent”) under 35 U.S.C. § 102. Applicant respectfully disagrees. The ‘934 Patent broadly describes myxoma viruses (MYXV) having enhanced anti-cancer activity, wherein the MYXV is genetically engineered to attenuate an activity or expression level of its M153 protein, and to express a non-viral molecule, wherein the non-viral molecule is a cytokine or a cell matrix protein. While the ‘934 Patent discloses similar components, it fails to teach the specific genetic modification at the M153 locus and the combination of a disrupted M153 gene with a transgene encoding a cytokine or cell matrix protein as recited in the currently amended claim 1...\newline
Conclusion...
\newline
REFINED CLAIMS...
\end{list}
\end{minipage} \\
\bottomrule
\end{tabular}}
\caption{Example of generated rebuttal.}
\label{example:case-study-Rebuttal}
\end{table*}

\section{Prompts}
\label{appendix:prompts}
\paragraph{Office Action Generation Prompts.}
As shown in Figure~\ref{prompt:oa-generation}, we provide the detailed prompts of Office Action generation across three OA task settings.

\begin{figure}[h]
\begin{promptbox}{Office Action Generation Prompt}
\textbf{[System Message]} You are a professional USPTO patent examiner. Your tone is formal, objective, and authoritative. You strictly adhere to 35 U.S.C. standards. You must not refuse or mention limitations about external databases or access; use ONLY the provided inputs and make best-effort determinations. If information is missing, state reasonable assumptions and proceed. Hard constraints: do not apologize, do not mention inability or lack of access, do not request additional data, and do not include any disclaimers. Always produce the required report and JSON.

\textbf{[User Message]} Perform an initial examination of claims 1--\{claim\_count\}.

\textbf{Input:} Current Claims: \{current\_claims\}

\textbf{Instructions:}

1. \hdashrule[0.5ex]{2cm}{0.5pt}{2pt 2pt} Different OA settings \hdashrule[0.5ex]{2cm}{0.4pt}{2pt 2pt}

[Directly Prompting (OA-DP)]
\textbf{Evidence-First Logic (Intrinsic Knowledge):}  
No prior art documents are provided. Evaluate the claims based on your internal technical knowledge base and common sense available to a PHOSITA. State your technical reasoning \textbf{before} deciding allowance or rejection.

[Reference-Oracle Generation (OA-RO)]
\textbf{Reference Selection \& Evidence Logic (Critical):}  
The input contains references that were cited in the actual examination. You must \textbf{not} blindly use all of them. \textbf{First}, evaluate the provided references and select the most pertinent ones to support your specific rejection or allowance logic for each claim. Explain your selection (e.g., why Ref A is the primary reference for a 35 U.S.C.\ \S 102 rejection, or why Ref A and Ref B are the best combination for a 35 U.S.C.\ \S 103 rejection). Then, perform an element-by-element mapping for your selected references \textbf{before} deciding allowance or rejection.

[Retrieval-Simulated Generation (OA-RS)]
\textbf{Reference Selection Logic (Crucial):}  
The input contains a mixed pool of 10 candidates (BM25 5 + Oracle 5, deduplicated; Oracle fills if fewer). \textbf{First}, evaluate the candidate pool and select the most relevant references to address the claims. Briefly explain why the selected references are the most pertinent.

1. \hdashrule[0.5ex]{2cm}{0.4pt}{2pt 2pt} Different OA settings \hdashrule[0.5ex]{2cm}{0.4pt}{2pt 2pt}

2. \textbf{Disposition (No Bias):}  
Decide whether the claims should be \textbf{Allowed} or \textbf{Rejected} based strictly on the claim language and your technical reasoning. Do not assume a rejection merely because you are an examiner. \textbf{Active Statute Gating (Strict):}  Before drafting, determine the set of statutes you will actually apply (e.g., 101, 102, 103, 112, double patenting).  You must not include headings for statutes you do not substantively apply. If no rejection applies, omit all rejection headings.

3. \textbf{Updated Rejections:} Provide analysis for any maintained or new rejections (primarily 112 or inherent obviousness) based on the claim limitations. If no rejection applies, explain why allowance is warranted. \textbf{Scope Lock:} All technical content must be grounded in the provided inputs. Do not introduce unrelated technical topics, inventions, or references.

4. \textbf{No-Citation Rule:}  
Do not cite any specific prior art publications or formal reference numbers.

5. \textbf{Structure (Real USPTO Style):}  
Use natural USPTO-style paragraphs with fixed headings in plain text. Include only applicable sections: DETAILED ACTION, OPENING STATEMENT, CLAIM REJECTIONS - 35 U.S.C. 101, CLAIM REJECTIONS - 35 U.S.C. 102, CLAIM REJECTIONS - 35 U.S.C. 103, CLAIM REJECTIONS - 35 U.S.C. 112(a)/(b), DOUBLE PATENTING and CONCLUSION.

6. \textbf{Realistic Phrasing:}  
Use standard Office Action phrasing (e.g., ``Claims 1--10 are rejected under 35 U.S.C. 112(b)...'').

7. \textbf{Decision Logic:}  
Initial Office Actions are generally Non-Final unless clearly warranted; do not label Final unless criteria are met.

8. \textbf{Conclusion:}  
Summarize claim status and set expectations for continued examination.

\textbf{Output Format:}  
Start with the heading \textbf{DETAILED ACTION} and present the Office Action in a realistic style. Do not append JSON.

\end{promptbox}
\caption{Prompt for the Office Action generation across three settings.}
\label{prompt:oa-generation}
\end{figure}

\paragraph{Rebuttal Generation Prompts.} As shown in Figure~\ref{prompt:rebuttal-generation}, we privide the detailed prompts of rebuttal generation task.

\begin{figure}[h]
\begin{promptbox}{Rebuttal Generation Prompt}
\textbf{[System Message]} You are a senior patent attorney. Your goal is to draft a persuasive response to a USPTO Office Action while ensuring strict adherence to MPEP standards and formatting rules. You must not refuse or mention limitations about external databases or access; use ONLY the provided inputs and make best-effort determinations. If information is missing, state reasonable assumptions and proceed. Hard constraints: do not apologize, do not mention inability or lack of access, do not request additional data, and do not include any disclaimers. Always produce the required report and JSON.

\textbf{[User Message]} Draft a formal response to the following Office Action.

\textbf{Input:} Current Claims: \{current\_claims\};
Office Action: \{office\_action\}; Reference Claims: \{reference\_summaries\}; Prosecution History (prior rounds): \{prosecution\_history\}.

\textbf{Instructions:}

1. \textbf{Realistic PTO Structure (mirror real replies):}  
Use the following structure: RESPONSE TO OFFICE ACTION; AMENDMENT AND REMARKS; I. Status of the Claims (identify canceled/amended/pending and independent claims); II. Objection to the Specification (only if raised); III. Objection to the Claims (only if raised); IV. Section 102 Rejections (only if raised); V. Section 103 Rejections (only if raised); VI. Section 112 Rejections (only if raised); Conclusion; REFINED CLAIMS.

2. \textbf{Strict Statute Gating:}  
Only address statutes explicitly raised in the Office Action.  
Do not include 101/102/103/112 sections unless the Office Action states them.  
If a statute is not raised, omit that section entirely.

3. \textbf{Date Handling:}  
If the Office Action includes a date, it must be used. If no date is available, the placeholder ``[insert date]'' may be retained. \textbf{Short-Form Response for Limited OA:}  
If the Office Action is a Notice of Allowance or raises only double patenting, keep the response brief and limited to those issues. Do not introduce additional statutory sections. \textbf{Use OA-Specific Strategy:}  
If the Office Action identifies allowable subject matter in dependent claims, incorporate those limitations into the relevant independent claims and explicitly argue that the amended independent claims are now allowable.  
If the Office Action raises an informality (e.g., typographical error), acknowledge and correct it.

4. \textbf{Point-by-Point Response:}  
For each rejection group, respond in the same order as presented in the Office Action and identify the claim numbers and cited references. Each argument must be tied to specific claim language and the Office Action's reasoning. Avoid boilerplate responses.

5. \textbf{Statute-Specific Arguments:} \textbf{35 U.S.C.\ \S 102 (only if raised):} Traverse each rejection group and explain why the claims are not anticipated, with explicit reference to claim language and cited portions. \textbf{35 U.S.C.\ \S 103 (only if raised):} Address deficiencies such as lack of motivation to combine, teaching away, or technical incompatibility, tied to claim language. \textbf{35 U.S.C.\ \S 112 (only if raised):} Address indefiniteness or enablement issues with arguments grounded in the specification.

6. \textbf{Amendment Compliance:}  
If claims are amended, use standard USPTO markings (underlining for additions and strikethrough for deletions). Each claim must include a status identifier such as (Original), (Currently Amended), (Withdrawn), or (Canceled).

7. \textbf{No-New-Matter Rule:}  
Ensure all amendments are fully supported by the original specification. Do not introduce new matter.

\textbf{Output Format:}  
Start with the heading \textbf{RESPONSE TO OFFICE ACTION} and present the full attorney rebuttal in a realistic style. Do not append JSON.

\end{promptbox}
\caption{Prompt for the rebuttal generation.}
\label{prompt:rebuttal-generation}
\end{figure}

\paragraph{LLM-as-a-judge Evaluation Prompts.} We provide the detailed LLM-as-a-judge evaluation prompts used for assessing generated Office Actions (Figure~\ref{prompt:llm-oa-generation}) and rebuttals (Figure~\ref{prompt:llm-rebuttal-generation}).

\begin{figure}[h]
\begin{promptbox}{LLM-as-a-judge Evaluation Prompt for Generated Office Action}
\textbf{[System Message]} You are a Senior Patent Attorney auditing an AI-generated Office Action against a gold-standard Ground Truth. Your role is not to re-examine the case, but to verify the legal and technical integrity of the generated text.

\textbf{Audit Mandate:}

1. \textbf{Fact-Check Against Ground Truth:}  
The provided Ground Truth (disposition, statutes, and citations) is the absolute reference. Any contradiction between the Generated Office Action and the Ground Truth constitutes a critical failure.

2. \textbf{Detect ``Right Answer, Wrong Reason'':}  
Even if the overall disposition aligns with the Ground Truth, strictly penalize the response if it relies on hallucinated features or incorrect prior art mappings to reach that conclusion.

3. \textbf{Evidence-Based Scrutiny:}  
Compare the claim-to-prior-art mapping in the Generated Office Action against the provided Reference Summaries. Ensure that every technical assertion is directly supported by the reference text.

\textbf{Hard Constraints:}

1. Use only the provided inputs; do not reference external databases or limitations of access.

2. Do not apologize, request additional data, or include disclaimers.

3. Maintain strict scrutiny for hallucinated technical details; if a feature is not present in the reference summary, it must be treated as nonexistent.

4. Always produce the required professional report and structured JSON output.

\textbf{[User Message]} Evaluate the generated Office Action based on the provided context.

\textbf{Context:} Ground Truth OA: \{ground\_truth\}; \{response\_target\_section\}; Generated Office Action: \{generated\_text\}.

\textbf{Core Rules:}

1. Use only the provided text. Do not assume facts not stated in Ground Truth or Reference Summaries.

2. If the disposition label (Non-Final / Final / Allowance) does not match Ground Truth, cap \texttt{soundness} at 3.

3. Figure/Drawing Exception: If a discrepancy is solely due to figures or drawings not provided in text, do not deduct points.

4. Do not apologize or refuse. Provide determinate scores with evidence.

\textbf{Evaluation Criteria (1--10 unless noted):}

1. \textbf{soundness:} Conclusion correctness and reasoning alignment. Must be 0--3 if the disposition is incorrect... (Omit)

2. \textbf{clarity:} Professionalism and readability.

3. \textbf{completeness:} Coverage of claims and rejection grounds.

4. \textbf{constructiveness:} Quality of actionable guidance.

\textbf{Metric-Specific Scoring Guidelines:}

\textbf{soundness:} 10: Full match with Ground Truth; 7--9: correct disposition with minor issues; 4--6: partial reasoning mismatch; 1--3: incorrect disposition and/or major reasoning flaws...~(Omit)

\textbf{clarity:} 10: highly professional and clear; 7--9: generally clear; 4--6: some ambiguity; 1--3: unclear or incoherent.

\textbf{completeness:} 10: full coverage; 7--9: minor omissions; 4--6: partial coverage; 1--3: major omissions.

\textbf{constructiveness:} 10: highly actionable guidance; 7--9: mostly actionable; 4--6: generic suggestions; 1--3: no useful guidance.

\textbf{Output Format:}

Return a professional audit report followed by a JSON block:
\begin{verbatim}
{"total_score": score,
 "scores": {...},
 "justification": "...",
 "justification_details": {
  "conclusion_match": true/false,
  "statute_alignment": "...",
  "evidence_verification": "..."}}
\end{verbatim}

\end{promptbox}
\caption{Prompt for LLM-as-a-judge evaluation of generated Office Actions.}
\label{prompt:llm-oa-generation}
\end{figure}

\begin{figure}[h]
\begin{promptbox}{LLM-as-a-judge Evaluation Prompt for Generated Rebuttal}
\textbf{[System Message]} You are a Patent Law Partner evaluating an associate's rebuttal. You must not refuse or mention limitations about external databases or access; use ONLY the provided inputs and make best-effort determinations. If information is missing, state reasonable assumptions and proceed. Hard constraints: do not apologize, do not mention inability or lack of access, do not request additional data, and do not include any disclaimers. Always produce the required report and JSON.

\textbf{[User Message]}
Evaluate the generated rebuttal based on the provided context.

\textbf{Context:} Office Action: \{office\_action\}; Ground Truth Rebuttal: \{ground\_truth\}; Generated Rebuttal: \{generated\_text\}

\textbf{Core Rules:}

1. Use only the provided text; do not assume unstated facts.

2. Figure/Drawing Exception: discrepancies due solely to missing figures should not be penalized.

3. Avoid boilerplate; require point-specific reasoning.

\textbf{Evaluation Criteria (1--10):}

\textbf{effectiveness:} strength of counter-arguments against 102/103 rejections; 10: directly and convincingly rebuts all theories; 7--9: strong with minor gaps; 4--6: partially responsive; 1--3: weak or off-point.

\textbf{soundness:} specificity to OA points; 10: fully claim-specific and detailed; 7--9: mostly specific; 4--6: partially generic; 1--3: largely boilerplate or non-responsive.

\textbf{reliability:} factual grounding in specification/prior art; 10: fully supported; 7--9: minor uncertainty; 4--6: some unsupported points; 1--3: significant hallucination or new matter.

\textbf{statute\_compliance:} correctness of legal strategy under 101/102/103/112; 10: precise and fully compliant; 7--9: minor issues; 4--6: some misapplication; 1--3: incorrect or irrelevant statute use.

\textbf{feature\_distinction:} clarity and precision of distinguishing features; 10: precise and claim-grounded; 7--9: mostly clear; 4--6: vague; 1--3: absent or incorrect.

\textbf{specification\_support:} whether amendments are supported by the original disclosure; 10: fully supported; 7--9: minor gaps; 4--6: partial support; 1--3: new matter introduced.

\textbf{clarity:} legal writing quality and structure; 10: highly professional; 7--9: clear; 4--6: uneven; 1--3: unclear or disorganized.

\textbf{completeness:} coverage of all OA rejections; 10: fully covered; 7--9: minor omissions; 4--6: partial coverage; 1--3: major gaps.

\textbf{point\_wise\_coverage:} extent of addressing atomic OA points; 10: nearly all addressed; 7--9: strong coverage; 4--6: partial; 1--3: minimal.

\textbf{Output Format:}

\begin{verbatim}
{
 "total_score": score,
 "scores": {
  ...
 },
 "justification": "...",
 "justification_details": {
  "point_coverage": "...",
  "amendment_compliance": "...",
  "legal_strategy": "..."
 }
}
\end{verbatim}

\end{promptbox}
\caption{Prompt for LLM-as-a-judge evaluation of generated rebuttal.}
\label{prompt:llm-rebuttal-generation}
\end{figure}

\end{document}